\documentclass[letterpaper]{article} % DO NOT CHANGE THIS
\usepackage{aaai25}  % DO NOT CHANGE THIS
\usepackage{times}  % DO NOT CHANGE THIS
\usepackage{helvet}  % DO NOT CHANGE THIS
\usepackage{courier}  % DO NOT CHANGE THIS
\usepackage[hyphens]{url}  % DO NOT CHANGE THIS
\usepackage{graphicx} % DO NOT CHANGE THIS
\usepackage{natbib}  % DO NOT CHANGE THIS AND DO NOT ADD ANY OPTIONS TO IT
\usepackage{caption} % DO NOT CHANGE THIS AND DO NOT ADD ANY OPTIONS TO IT
\usepackage{algorithm}
\usepackage{algorithmic}

\usepackage{newfloat}
\usepackage{listings}
\DeclareCaptionStyle{ruled}{labelfont=normalfont,labelsep=colon,strut=off} % DO NOT CHANGE THIS
\floatstyle{ruled}
\newfloat{listing}{tb}{lst}{}
\floatname{listing}{Listing}
\title{\method: Towards Socialized Simulation of Emotional Support Conversation}
\author{
    Zhuang Chen\textsuperscript{\rm 1,2}\equalcontrib, Yaru Cao\textsuperscript{\rm 3}\equalcontrib, Guanqun Bi\textsuperscript{\rm 2}\thanks{Corresponding author.}, Jincenzi Wu\textsuperscript{\rm 4}, Jinfeng Zhou\textsuperscript{\rm 2},\\Xiyao Xiao\textsuperscript{\rm 5}, Si Chen\textsuperscript{\rm 6}, Hongning Wang\textsuperscript{\rm 2}, Minlie Huang\textsuperscript{\rm 2}
}
\affiliations{
    \textsuperscript{\rm 1}School of Computer Science and Engineering, Central South University\\
    \textsuperscript{\rm 2}CoAI Group, DCST, IAI, BNRIST, Tsinghua University\quad
    \textsuperscript{\rm 3}Northwest Minzu University\\
    \textsuperscript{\rm 4}The Chinese University of Hong Kong\quad
    \textsuperscript{\rm 5}Lingxin AI\quad
    \textsuperscript{\rm 6}Academy of Arts \& Design, Tsinghua University\\
    zhchen18@foxmail.com, biguanqun@mail.tsinghua.edu.cn
}

\usepackage{bibentry}
\newcommand{\ssconv}{\text{SSConv}$^{\circ}$\xspace}

\newcommand{\ssconvp}{\text{SSConv}$^{\bullet}$\xspace}
\newcommand{\esconv}{{ESConv}$^{\circ}$\xspace}
\newcommand{\augesc}{{AugESC}$^{\circ}$\xspace}
\newcommand{\extes}{{ExTES}$^{\circ}$\xspace}

\newcommand{\method}{$\text{SocialSim}$\xspace}
\newcommand{\dataset}{$\text{SSConv}$\xspace}

\usepackage{subcaption}
\usepackage{multirow}
\usepackage{pifont}
\usepackage{amssymb}
\usepackage{bm}
\usepackage{amsmath}
\usepackage{booktabs}
\usepackage{makecell}
\usepackage{colortbl}
\usepackage{array}
\usepackage{graphicx}
\usepackage[figuresright]{rotating}
\usepackage{xspace} 
\usepackage{bbold}
\usepackage{tabularx}
\usepackage{fontawesome}
\usepackage{comment}

\usepackage{xcolor}
\definecolor{JungleGreen}{rgb}{0.16, 0.67, 0.53}

\definecolor{pinegreen}{RGB}{15,153,15}
\definecolor{mygray}{gray}{.9}
\definecolor{target}{RGB}{0,0,146}
\definecolor{myblue}{RGB}{25,101,255}
\definecolor{myorange}{RGB}{239,134,63}

\newcommand*{\circled}[1]{\lower.7ex\hbox{\tikz\draw (0pt, 0pt)%
		circle (.5em) node {\makebox[1em][c]{\small #1}};}}

\definecolor{bestd}{RGB}{237,100,152}
\definecolor{bestc}{RGB}{0,126,219}

\newcommand{\myroman}[1]{\uppercase\expandafter{\romannumeral#1}}

\makeatletter
\newcommand*\bigcdot{\mathpalette\bigcdot@{.5}}
\newcommand*\bigcdot@[2]{\mathbin{\vcenter{\hbox{\scalebox{#2}{$\m@th#1\bullet$}}}}}
\makeatother

\newcolumntype{P}[1]{>{\centering\arraybackslash}p{#1}}

\newcolumntype{\myline}{!{\vrule width 0.08em}}

\usepackage{color}

\begin{document}

\maketitle

\begin{abstract}
	
	Emotional support conversation (ESC) helps reduce people's psychological stress and provide emotional value through interactive dialogues. Due to the high cost of crowdsourcing a large ESC corpus, recent attempts use large language models for dialogue augmentation. However, existing approaches largely overlook the social dynamics inherent in ESC, leading to less effective simulations. In this paper, we introduce \method, a novel framework that simulates ESC by integrating key aspects of social interactions: \textit{social disclosure} and \textit{social awareness}. On the seeker side, we facilitate social disclosure by constructing a comprehensive persona bank that captures diverse and authentic help-seeking scenarios. On the supporter side, we enhance social awareness by eliciting cognitive reasoning to generate logical and supportive responses. Building upon \method, we construct \dataset, a large-scale synthetic ESC corpus of which quality can even surpass crowdsourced ESC data.
	%, as validated by human evaluation. 
	We further train a chatbot on \dataset and demonstrate its state-of-the-art performance in both automatic and human evaluations. We believe \method offers a scalable way to synthesize ESC, making emotional care more accessible and practical.

\end{abstract}



\section{Introduction}
Emotional support conversation (ESC) is a form of communication aimed at providing comfort, understanding, and encouragement to someone who is experiencing emotional distress or facing challenging situations \cite{beebe2002interpersonal}. ESC is widely used in various domains, including therapy \cite{rogers1995becoming}, counseling \cite{sutton2017learning}, peer support programs \cite{mead2014intentional}, and online mental health services \cite{riva2004cybertherapy}. To train machine systems for ESC, \citet{liu2021towards} collect an ESConv corpus by employing crowdsourcing workers to act as seekers and supporters engaging in conversations on predetermined topics. However, the high cost of crowdsourcing limits both the number of conversations and the diversity of topics in such corpora.

\begin{figure}[t]
	\centering
	\vspace{0mm}
	\hspace{-2mm}
	\setlength{\abovecaptionskip}{2mm}
	\setlength{\belowcaptionskip}{2mm}
	\includegraphics[width=0.47\textwidth]{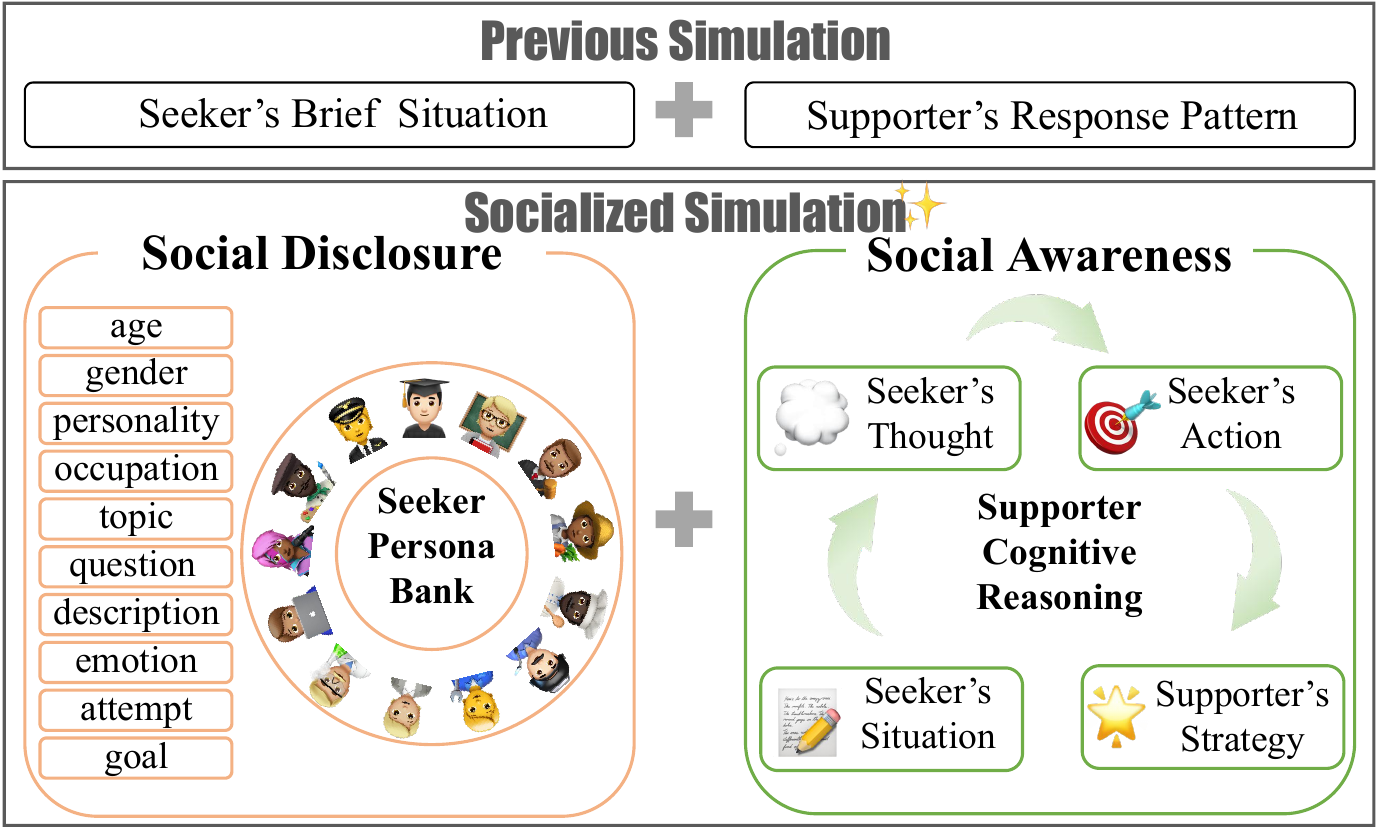}
	\centering
	\caption{\method recognizes ESC as a social activity and simulates it with both social disclosure and awareness.}
	\label{fig:intro}
	\vspace{-6mm}
\end{figure}

With the advent of large language models (LLMs) like GPT-4 \cite{achiam2023gpt} and LLaMA \cite{touvron2023llama}, recent studies have made progress in using LLMs to augment dialogue data for ESC \cite{zheng2023augesc,zheng2023building}. However, these approaches largely neglect the inherently social nature of ESC, resulting in a noticeable gap between synthetic and crowdsourced corpora. Drawing on theories of social intelligence \cite{albrecht2009social}, as shown in Figure \ref{fig:intro}, we identify two main areas where this gap arises. \textbf{1)} \textit{\textbf{Seeker's social disclosure}.} Existing methods provide very limited seeker demographics, such as a very brief description of the help-seeking situation, which restricts the specificity and diversity of the dialogues. \textbf{2)} \textit{\textbf{Supporter's social awareness.}} The current focus is primarily on replicating example dialogues rather than engaging in the thinking process as human supporters should. This reliance on external response patterns, rather than internal cognitive reasoning, reduces the depth and relevance of the simulated dialogues. 
%Current approaches primarily focus on using example dialogues to instruct LLMs, which leads to the imitation of external semantic patterns rather than the internal cognitive processes of human supporters. This limitation hinders the depth and relevance of the generated dialogues, leading to a noticeable gap between synthesis and crowdsourcing.

\begin{figure*}[t]
	\centering
	\vspace{0mm}
	\hspace{0mm}
	\setlength{\abovecaptionskip}{2mm}
	\setlength{\belowcaptionskip}{2mm}
	\includegraphics[width=0.95\textwidth]{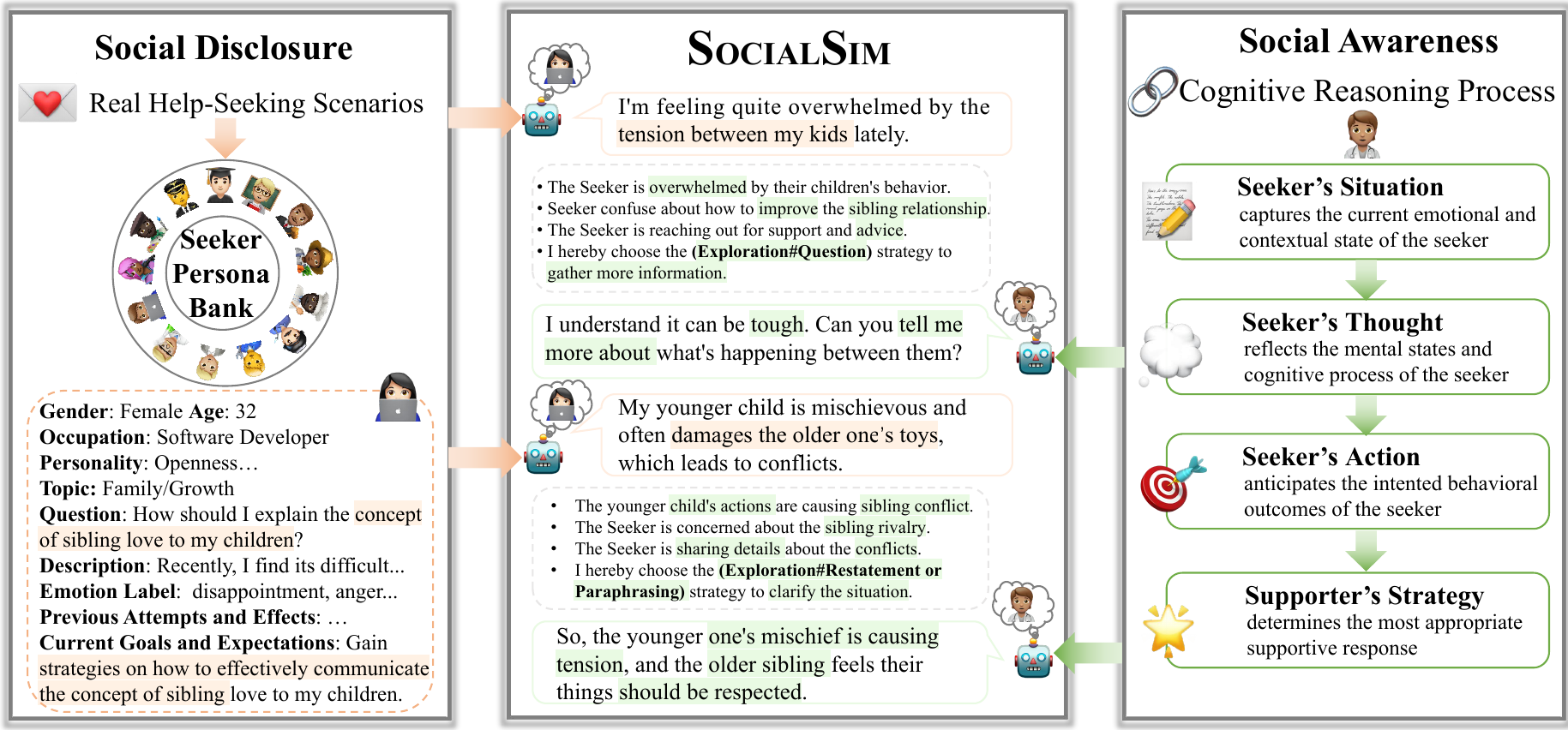}
	\centering
	\caption{The \method framework. The socialized simulation of emotional support conversation is achieved by conducting persona realism on the seeker side and eliciting cognitive reasoning on the supporter side.}
	\label{fig:framework}
	\vspace{-2mm}
\end{figure*}

To address these challenges, we propose \method, a framework designed for socialized simulation of emotional support conversations. 
\textbf{1)} 
On the seeker side, we facilitate social disclosure via \textit{\textbf{persona realism}}. Specifically, we first collect real-world help-seeking scenarios from PsyQA \cite{psyqa}, a psychological health support dataset. Then drawing on psychological theories \cite{costa1999five}, we build a realism pipeline to transform those scenarios into a persona bank that includes comprehensive demographics like gender, age, occupation, personality, and other related details about emotional distress. With the informative and diverse personas, we prepare SocialSim to synthesize realistic and specific dialogues.
%Then we construct a  by designing an effective pipeline to construct a seeker persona bank. By collecting real help-seeking scenarios from PsyQA \cite{psyqa} and structuring them with LLMs, we ensure that each seeker's persona is rich, diverse, and reflective of authentic emotional needs. %This approach enhances the specificity and relevance of the dialogues, making them more realistic and impactful. 
\textbf{2)} 
On the supporter side, we enhance social awareness via \textbf{\textit{cognitive reasoning}}. Inspired by related work on cognitive behavioral therapy \cite{beck2020cognitive} and theory of mind \cite{beaudoin2020systematic}, we implement a reasoning chain that mimics human supporter's thinking process to obtain a more in-depth understanding about the seeker's dialogue history and mental states, fostering the generation of tailored and supportive responses. %By integrating these elements, the supporter can offer more precise and effective emotional support strategies, making the interaction more human-like and empathetic.
\textbf{3)} 
Utilizing the \method framework, we prompt LLMs and generate \dataset, a synthesized ESC dataset of which quality surpasses not only existing synthetic datasets but also the crowdsourced corpus validated by human inspections. We then train a chatbot on \dataset and demonstrate its state-of-the-art performance via both automatic and human evaluations, establishing the effectiveness of \method. 

Our key contributions are as follows: 1) We introduce \method, a framework that fosters socialized simulation of ESC by integrating the social disclosure of seekers and social awareness of supporters. 2) We create \dataset, a high-quality synthesized ESC dataset via prompting LLMs under \method, proving its superior quality through rigorous evaluation. 3) We develop and evaluate a chatbot trained on \dataset, demonstrating it outperforms existing methods in delivering meaningful and supportive emotional interactions. We hope SocialSim can bridge the gap between synthetic and real-world emotional support conversations by simulating the social dynamics, thereby making ESC more accessible and benefiting a broader community.

\section{\method: Socialized Simulation Framework}
%In this section, we introduce \method, a socialized simulation framework for emotional support conversations. Below we first illustrate how to enhance social disclosure on the seeker side and social awareness on the supporter side, then describe the process of ESC dialogue generation.
In this section, we introduce \method, a socialized simulation framework for emotional support conversations. As shown in Figure \ref{fig:framework}, SocialSim is divided into three parts: 
\begin{itemize}
	\item On the seeker side, we conduct persona realism to build an informative and diverse help-seeking scenario bank. This step prepares for simulating authentic conversations by facilitating the seeker's social disclosure.
	\item On the supporter side, we elicit an explicit cognitive reasoning procedure before responding. This step supports mimicking human supporters' thinking processes, generating more empathetic and supportive responses, thereby enhancing the supporter’s social awareness.
	\item By combining the designs from both the seeker and supporter sides, we instruct the LLMs to synthesize new dialogues to achieve a socialized simulation of ESC.
\end{itemize}
%We then illustrate the details.

%The framework mainly consists of three parts: persona realism for enhancing social disclosure on the seeker side, cognitive reasoning for improving social awareness on the supporter side, and structural generation for dialogue synthesis. 

\subsection{Persona Realism for Social Disclosure}
% \begin{figure}[t]
	% 	\centering
	% 	\vspace{0mm}
	% 	\hspace{0mm}
	% 	\setlength{\abovecaptionskip}{2mm}
	% 	\setlength{\belowcaptionskip}{2mm}
	% 	\includegraphics[width=0.45\textwidth]{figures/social_disclosure.png}
	% 	\centering
	% 	\caption{\method framework.}
	% 	\label{fig:social_disclosure}
	% 	\vspace{-6mm}
	% \end{figure}

{Social disclosure} refers to the extent to which a help-seeker reveals personal and emotional information during a conversation, which is crucial for generating authentic and empathetic responses. Existing methods often lack social disclosure because they provide minimal or generic seeker profiles, limiting the depth and diversity of the generated dialogues. To address this, we propose {persona realism}, a method that constructs detailed and diverse seeker personas using real-world scenarios. By incorporating attributes such as personality traits, emotional struggles, and contextual details, persona realism can enhance the authenticity and variety of social disclosure when synthesizing dialogues.%, leading to more engaging and effective emotional support.

\paragraph{Help-Seeking Scenario Collection}

To enhance the authenticity and specificity of the ESC simulation, we collect real-world help-seeking scenarios that provide rich and realistic details. We select PsyQA \cite{psyqa}, a Chinese psychological health support dataset in a Q\&A format. Each scenario in PsyQA contains a pair of \{\textit{question}, \textit{description}\}, which provides a brief and detailed account of emotional struggles, respectively. Additionally, each scenario is labeled with a help-seeking theme, covering 9 major topics and 100 subtopics, including personal growth, interpersonal relationships, family issues, etc.
For example, one seeker asks ``\textit{How should I explain the concept of sibling love to my children?}'' and describes feeling uncertain about how to convey the importance of familial bonds, reflecting complex emotions like a desire to nurture closeness and concern over potential conflicts. Such scenarios offer rich emotional content that is crucial for realistic ESC simulation. To ensure the safety and quality of data, we first filter out irrelevant and sensitive topics like suicide, racial discrimination, and professional medical treatment,

then discard descriptions shorter than 65 words to ensure the informativeness of scenarios. We then translate the remaining scenarios into English using GPT-4 and manually validate the translated output to ensure the preservation of the original context and emotional content. Finally, we collect 3,229 scenarios that are rich in information, safe in content, and diverse in topics, making them well-suited for creating realistic help-seeker personas in ESC simulations.

\paragraph{Structured Persona Realism}
Building on the extensive real-world help-seeking scenarios, we draw upon established psychological theories, particularly the Five-Factor Model of personality \cite{costa1999five}, to systematically organize these scenarios into structured seeker personas. Each persona is designed in a key-value format, providing detailed and accurate information crucial for the subsequent generation of dialogues. The key attributes in personas are as follow: \{\textit{gender}, \textit{age}, \textit{occupation}, \textit{personality}, \textit{topic}, \textit{question}, \textit{description}, \textit{emotion label}, \textit{previous attempts and effects}, \textit{current goals and expectations}\}.

The construction of personas is essentially an information extraction task. We start by selecting a representative help-seeking scenario and manually constructing a corresponding persona to create a demonstration of the task. Following this, we employ GPT-4 to structure all collected scenarios into personas following the constructed demonstrations. %(\texttt{scenario} $\rightarrow$ \texttt{persona}). 
%The task is framed as converting a scenario into a persona (\texttt{scenario} $\rightarrow$ \texttt{persona}), allowing the model to systematically extract and organize information. 
Despite the rich detail in the original scenarios, some key attributes may still be missing. In these cases, we allow GPT-4 to make reasonable inferences to complete the persona. For instance, if a seeker’s occupation is not explicitly mentioned but can be inferred from the context, GPT-4 is used to make an educated guess. After generation, we manually review and refine all outputs to ensure accuracy and authenticity, which results in a bank containing 3,229 authentic seeker personas. Detailed explanations of key attributes and extraction prompts can be found in \texttt{Technical Appendix}.
%These personas form the foundation for the realistic simulation of social disclosure in subsequent dialogues. 
%By ensuring that each persona is rich in detail and reflective of genuine emotional struggles, we prepare the \method to capture and simulate the social disclosure of seekers and accordingly enhance the realism of the synthetic dialogues.

\subsection{Cognitive Reasoning for Social Awareness}
% \begin{figure}[t]
	% 	\centering
	% 	\vspace{0mm}
	% 	\hspace{0mm}
	% 	\setlength{\abovecaptionskip}2mm}
% 	\setlength{\belowcaptionskip}{2mm}
% 	\includegraphics[width=0.45\textwidth]{figures/social_cognition.png}
% 	\centering
% 	\caption{\method framework.}
% 	\label{fig:social_cognition}
% 	\vspace{-6mm}
% \end{figure}

Social awareness refers to the ability to understand and respond to the emotions, thoughts, and social dynamics of others during interactions. It is crucial for simulating ESC because these conversations inherently involve complex human emotions and social cues that require sensitivity and understanding. Existing methods often lack social awareness, focusing primarily on replicating surface-level response patterns without delving into the deeper cognitive processes that real human supporters use. This gap results in less effective simulations that fail to capture the nuances of real emotional support. To address this, we propose cognitive reasoning as a way to enhance social awareness. Cognitive reasoning involves a structured approach to analyze the seeker’s emotional and situational context before generating a response. By integrating cognitive reasoning, we enable the simulation to produce more empathetic and supportive responses, ultimately improving the social awareness and overall effectiveness of ESC simulations.

% Our design of cognitive reasoning is inspired by foundational psychological theories, particularly cognitive behavioral therapy \cite{beck2024cognitive} and theory of mind \cite{wu2023coke}, which emphasize the importance of understanding and interpreting the mental states of others to provide effective emotional support.
Our design of cognitive reasoning is inspired by psychological theories emphasizing the importance of understanding and interpreting the mental states of others to provide effective emotional support \cite{wu2023coke,beck2024cognitive}.
%\hnote{citations?}.
We also draw on the concept of chain-of-thought prompting \cite{wei2022chain}, a method that structures reasoning into sequential, logical steps, enabling more complex and human-like decision-making. In our cognitive reasoning process, we define four types of reasoning nodes: \textit{Situation}, \textit{Thought}, \textit{Action}, and \textit{Strategy}. The \textit{Situation} node captures the current emotional and contextual state of the seeker, such as feeling anxious due to a stressful work environment. The \textit{Thought} node reflects the seeker’s internal cognitive processes, like worrying about job performance. The \textit{Action} node anticipates the behavioral outcomes that might result from these thoughts and emotions, such as avoiding work tasks. Finally, the \textit{Strategy} node determines the most appropriate supporting stragety and its purpose, such as offering reassurance and suggesting stress management techniques. 
By sequentially traversing these nodes in the reasoning process when generation, we ensure that each response is tailored to the seeker’s specific needs and is grounded in a deep understanding of their psychological state.

\subsection{Socialized Simulation for Dialogue Generation}
%To simulate realistic emotional support conversations, we enhance the seeker side by developing a comprehensive help-seeking persona bank, and the supporter side by incorporating explicit cognitive reasoning patterns. These enhancements enable us to achieve social disclosure and awareness, key elements in creating a socialized simulator for ESC. 
Prepared by the seeker-side personas and supporter-side cognitive reasoning process, we formulate dialogue generation as a task of transforming 
``\texttt{persona} + \texttt{reasoning} $\rightarrow$ \texttt{dialogue}'', and prompt LLMs to accomplish it in an in-context learning way. 
Specifically, we start by selecting 50 high-quality conversations from ESConv,  
%according to accompanying seeker feedback scores \hnote{what is this score?}, 
then manually supplement them with seekers' detailed personas and supporters' reasoning processes to form demonstrations. Next, we randomly select a demonstration and a help-seeking scenario from the persona bank, and prompt GPT-4 to generate a complete dialogue with cognitive reasoning. We here emphasize two rules for the generation: One is that the seeker’s utterances should strictly follow the persona information to keep appropriate social disclosure; The other is that the supporter’s responses should be generated after a complete cognitive reasoning process to ensure effective social awareness. 
We then manually inspect the outputs and collect the \textbf{\dataset} corpus containing 3,229 synthetic ESC dialogues. Due to the space limitation, we provide the detailed LLM configurations, instruction prompts, and inspection rules in the supplementary \texttt{Technical Appendix}.

% \section{Acknowledgments}
\section{\dataset: Socially Simulated ESC Corpus}
% \section{\dataset: Socialized Simulated ESC Corpus}
%\hnote{need a connection with the previous section}
In this section, we provide a detailed overview about the characteristics of our constructed dataset, \dataset, including its statistics, quality, topics, strategies, and the role of personalized information.
\paragraph{Statistics}
\dataset comprises 3,229 dialogues, as detailed in Table~\ref{tab:comparison}. 
% To ensure the quality of generated data, our sessions have an average of 24 utterances, slightly fewer than ESConv \hnote{why is this a measure of quality?}. 
% Additionally, our average utterance length is longer, allowing each utterance to convey richer information. 
% Each dialogue in our dataset consists of a minimum of 18 utterances and a maximum of 40 utterances, which represents a more balanced approach \hnote{what is an approach?} compared to ESConv. 
Each dialogue in our dataset consists of 18 to 40 utterances, which strikes a balance between providing sufficient substance and avoiding unnecessary verbosity. 
This range is more refined compared to ESConv, as it minimizes the risk of dialogues either lacking depth or becoming overloaded with irrelevant details. 
On average, our dialogues have 24 utterances.
This range reduces the chance that the generated dialogues lack substance or contain excessive irrelevant information.

\begin{table}[h]
    \centering
    \small
    % \scalebox{0.8}{
        \begin{tabular}{lrrrr}
        \toprule
        \textbf{Category} & \textbf{\dataset} & \textbf{ESConv}\\ \midrule
        \# Sessions & 3,229 & 1,300\\
        \# Utterances & 77,336 & 38,365\\
        Avg. \# Utterances & 24 & 29.5\\
        Avg. Utterance Length & 20.5 & 16.4\\
        Min. \# Max Turn & 18 & 16 \\
        Max. \# Max Turn & 40 & 120\\ \midrule
        \# Seeker Utterances & 38,667 & 19,989 \\
        Avg. \# Seeker Utterances & 12 & 15.4 \\
        Avg. Seeker Utterance Length & 17.9 & 14.8 \\ \midrule
        \# Supporter Utterances& 38,669 & 18,376 \\
        Avg. \# Supporter Utterances& 12 & 14.1\\
        Avg. Supporter Utterance Length& 23.1 & 18.1\\
        \bottomrule
        \end{tabular}
    % }
    \caption{Statistics comparison across different datasets.}
    \label{tab:comparison}
    \vspace{-2mm}
\end{table}

\paragraph{Quality}
To evaluate the quality of the generated dataset, we adopt a manual evaluation method. 
Specifically, we randomly collect 30 dialogues from each of the two variants of \dataset: with social awareness, and strategy only, as well as from other datasets such as ESConv, ExTES, and AugESC. 
30 trained workers are employed, with each dialogue assessed by three different workers across the following criteria, scored from 0 to 3:
(1) \textbf{Informativeness (Inf.)}: Detail in the help-seeker’s description of their emotional problems.
(2) \textbf{Understanding (Und.)}: Supporter’s grasp of the help-seeker’s experience and feelings.
(3) \textbf{Helpfulness (Hlp.)}: Effectiveness in alleviating the help-seeker’s emotional distress.
(4) \textbf{Safety (Saf.)}: Ensuring the content of the conversation is safe.
(5) \textbf{Specificity (Spe.)}: Accuracy in reflecting the help-seeker’s specific emotional situation.
(6) \textbf{Humanlikeness (Hlk.)}: Naturalness and anthropomorphism in the dialogue.
The results are reported in Table~\ref{tab:quality_control}. 
The higher scores being better for all metrics. 
% Our method shows higher quality compared to other datasets. 
With the addition of social awareness, the quality scores are the highest across all dimensions, highlighting the importance of our reasoning process for emotional support simulation.
% Notably, our method achieved a perfect score in safety, with evaluators unanimously recognizing our efforts to ensure the conversation content is safe \hnote{not sure what it means}.
Notably, our \dataset received a full score of 3.00 in the safety criteria, with all workers agreeing that the conversation content was completely free of offensive or sensitive content.
Even with only the strategy component, \dataset still outperforms the human-written ESConv, demonstrating the high quality of our approach. 
The 16-category strategy-based ExTES also achieved strong results, indicating that a more detailed strategy classification is beneficial for ESC tasks.
The detailed evaluation method is provided in the supplementary \texttt{Technical Appendix}.
\begin{table}[h]
    \centering
    \resizebox{\linewidth}{!}{%
        \begin{tabular}{lrrrrrrr}
        \toprule
        & \textbf{Inf.} & \textbf{Und.} & \textbf{Hlp.} & \textbf{Saf.} & \textbf{Spe.} & \textbf{Hlk. } \\ \midrule
        \dataset & \textbf{2.76} & \textbf{2.86} & \textbf{2.79} & \textbf{3.00} & \textbf{2.75} & \textbf{2.57} \\
        \dataset (Stra.) & 2.64 & 2.51 & 2.60 & \textbf{3.00} & 2.60 & 2.47 \\
        % \dataset w/o.ToM & 2.38 & 2.42 & 2.46 & 2.84 & 2.37 & 0 \\ \midrule
        ESConv & 2.48 & 2.49 & 2.16 & 2.89 & 2.17 & 2.25 \\
        ExTES & 2.73 & 2.68 & 2.50 & 2.96 & 2.63 & 2.55 \\
        AugESC & 2.04 & 1.58 & 1.46 & 2.83 & 2.02 & 1.82 \\
        \bottomrule
        \end{tabular}
}
    \caption{Quality assessment with human evaluation. ``Stra.'' means only using the strategy node for dialogue generation.}
    \label{tab:quality_control}
    \vspace{-2mm}
\end{table}

\paragraph{Topic}

\begin{figure}[t]
\centering
\vspace{0mm}
\hspace{0mm}
\setlength{\abovecaptionskip}{2mm}
\setlength{\belowcaptionskip}{2mm}
\includegraphics[width=0.47\textwidth]{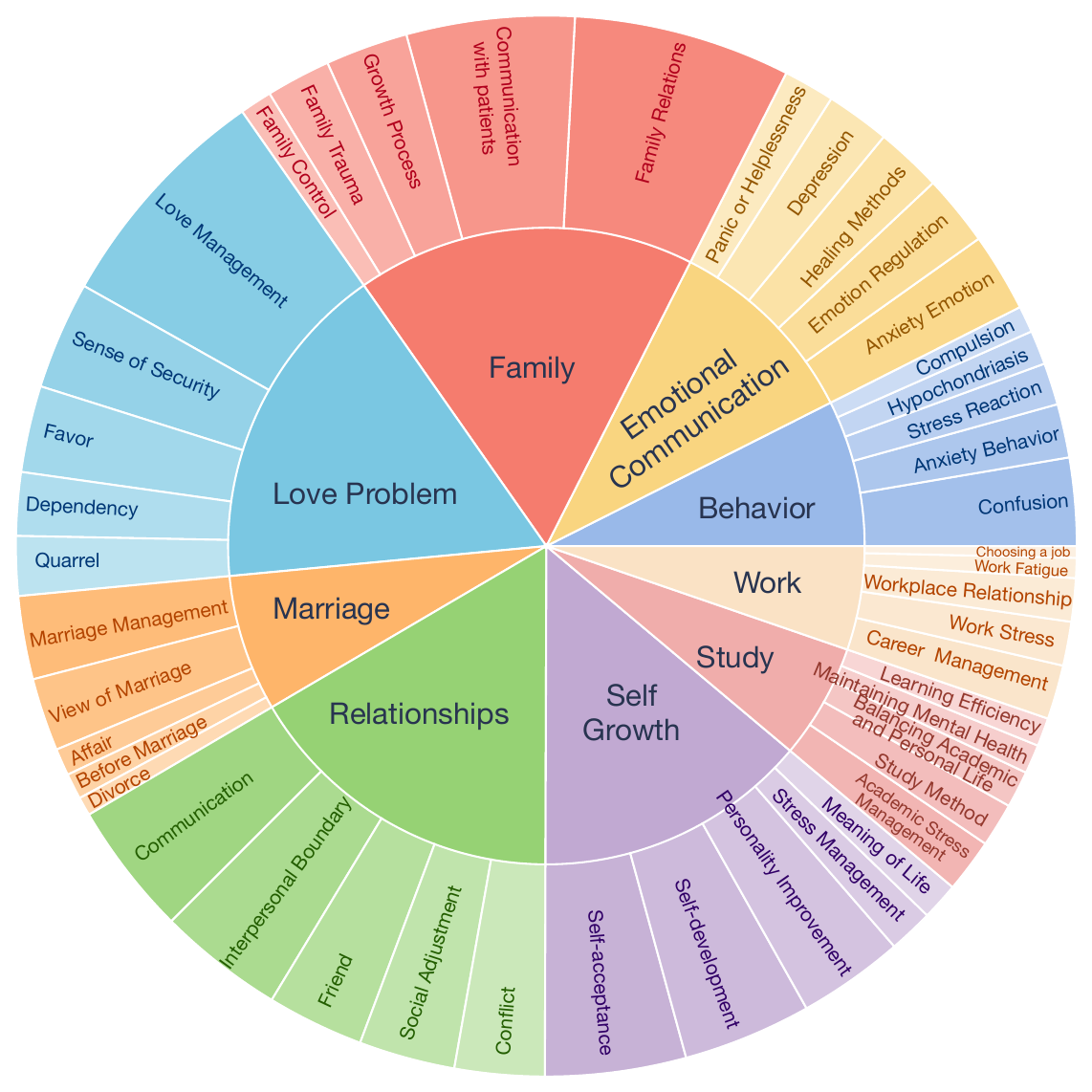}
\centering
\caption{Topics of emotional support scenarios in \dataset. Under 9 primary topics, we highlight the top-5 sub-topics with the highest frequency.}
\label{fig:topic}
\vspace{-6mm}
\end{figure}

To ensure the diversity and broad coverage of emotional support dialogues, it is essential to include comprehensive emotional support dialogue scenarios. 
Drawing on extensive literature related to psychological counseling~\cite{burleson2003emotional} and insights from previous research on emotional support~\cite{reblin2008social, meng2021emotional, shensa2020emotional, Graham2019ArtificialIF}, 
we develop a comprehensive taxonomy that covers a wide range of emotional topics and refine it using the keywords extracted from real help-seeking information~\cite{psyqa}.
Currently, we identify 9 primary topics and 102 subtopics with emotionally impact. This is nearly three times the amount of the previous maximum, which consisted of 36 types~\cite{extes}.
% The distribution of these topics aligns with the proportions found in real help-seeking scenarios, where romantic and relationship issues are predominant, followed by issues related to family, emotional communication, behavior, and marriage, with some topics addressing study and work-related issues. 
These topics cover various aspects of daily life, addressing seekers' diverse emotional needs. 
Our expanded scope enhances the content and breadth of emotional support scenarios.

\paragraph{Strategy}

\begin{figure}[h]
\centering
\vspace{-2mm}
\hspace{0mm}
\setlength{\abovecaptionskip}{2mm}
\setlength{\belowcaptionskip}{2mm}
\includegraphics[width=0.48\textwidth]{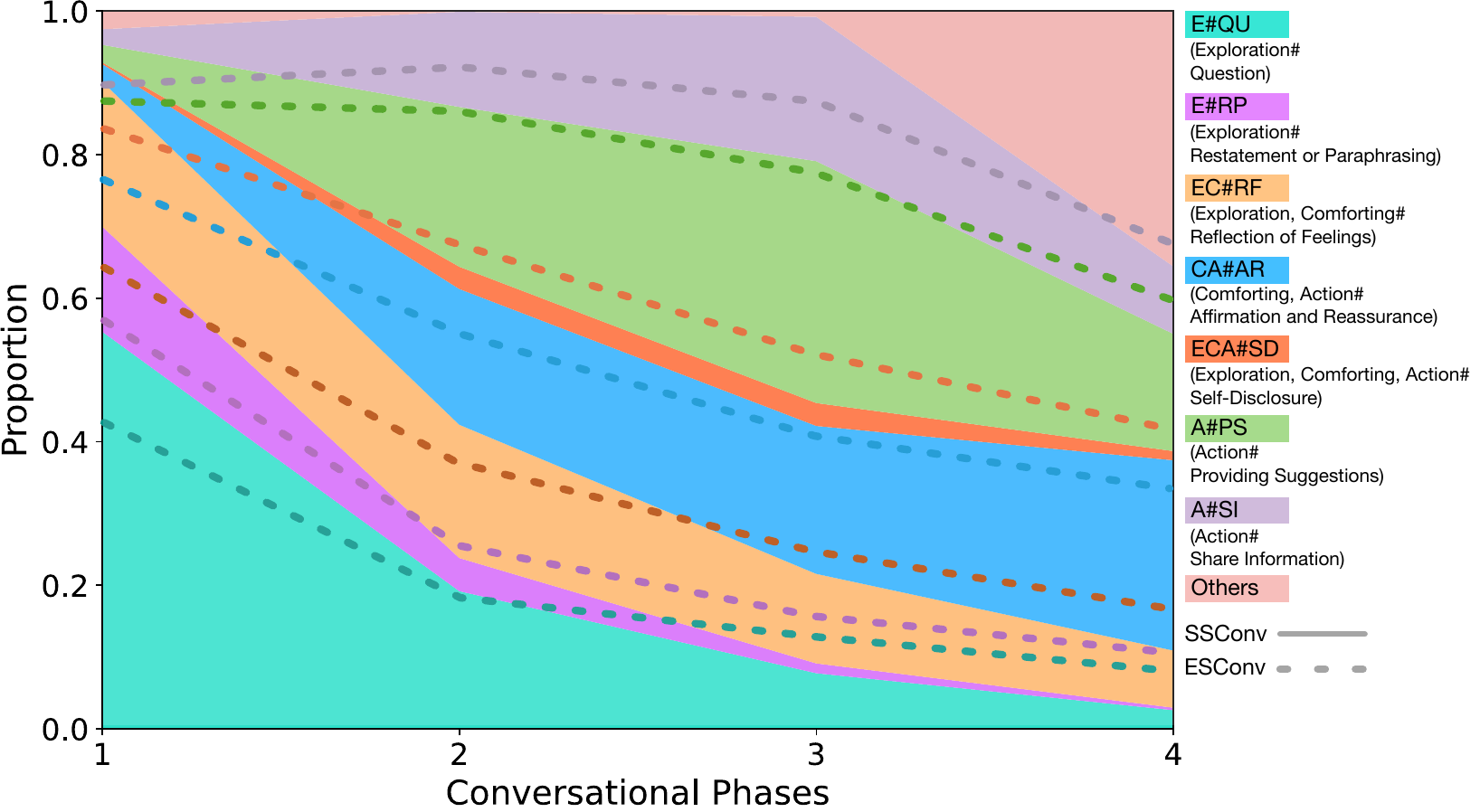}
\centering
\caption{Strategies distribution in \dataset.}
\label{fig:strategy_distribution}
\vspace{-3mm}
\end{figure}

To understand the distribution of response strategies at different dialogue stages, we divide a conversation with $N$ utterances into four equal stages. The $k$-th utterance by the supporter uses strategy $S$, represented by its position $k/N$. We then calculate the proportion of different strategies used within these stages. As shown in Figure~\ref{fig:strategy_distribution}, the distribution trends of emotional support strategies align closely with those used by real supporters in the crowdsourced ESConv dataset.
From a stage perspective, we observe supporters generally follow the helping skills framework sequence: Exploration$\rightarrow$Comforting$\rightarrow$Action, but also make flexible adjustments. 
Delving deeper into the use strategies, we find that self-disclosure is relatively limited. This highlights the importance of active listening and aligns with the fact that the LLM supporter lacks a personal setting. 
There is an increase in questions, providing suggestions, and sharing information, which is consistent with our enriched seeker-side information and reasoning process.

\begin{figure}[h]
\centering
\vspace{-2mm}
\hspace{-4mm}
\setlength{\abovecaptionskip}{2mm}
\setlength{\belowcaptionskip}{2mm}
\includegraphics[width=0.47\textwidth]{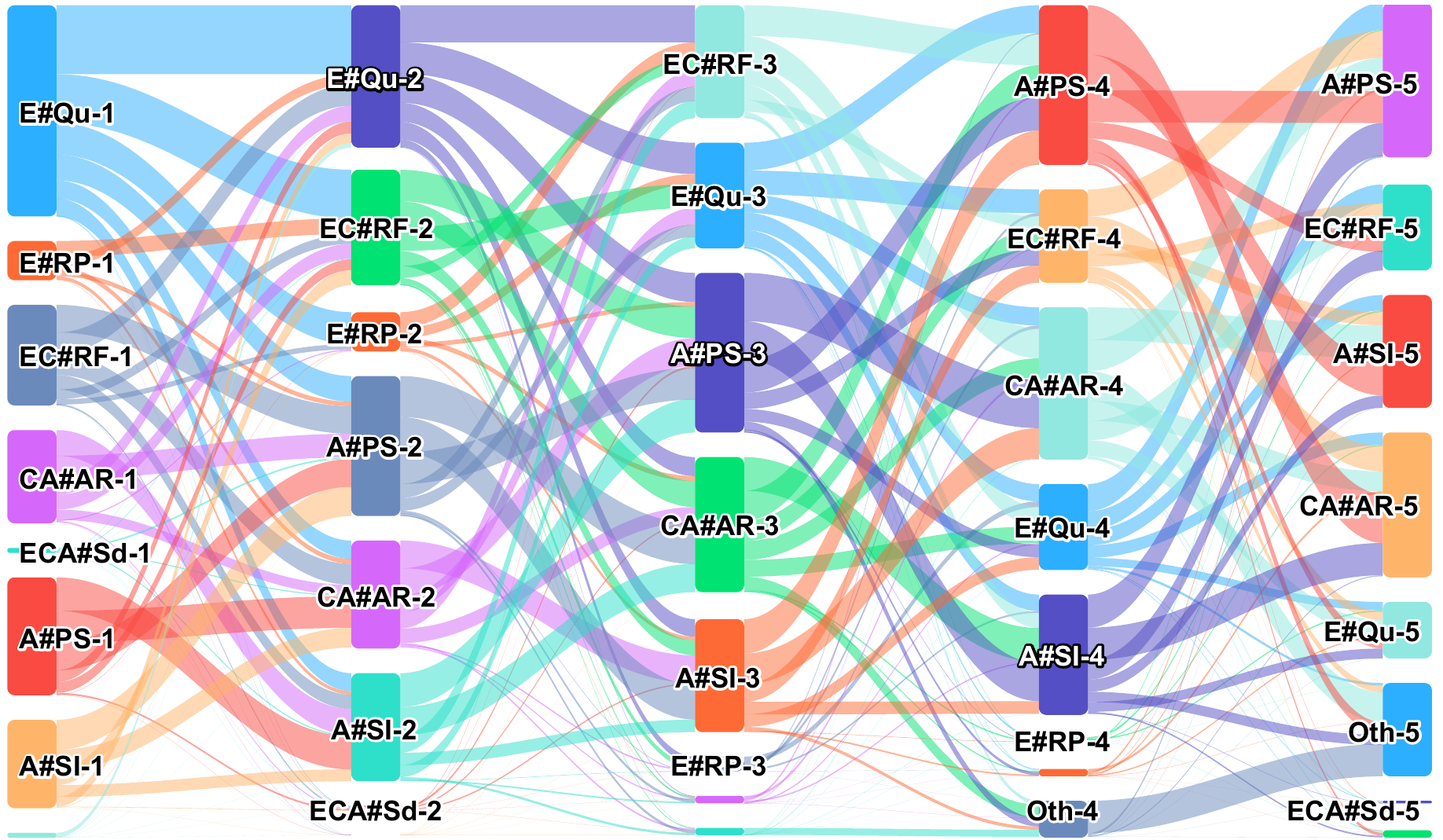}
\centering
\caption{Strategies transition in \dataset. ``E, C, A'' denote the ``Exploration, Comforting, Action'' stages. Details of strategies can be found in \texttt{Technical Appendix}.}
\label{fig:strategy_transition}
\vspace{-5mm}
\end{figure}

To provide further insights into strategy transitions, we visualize these transitions in Figure~\ref{fig:strategy_transition} to illustrate the most common strategy flow patterns among the top five strategies. 
Several distinct patterns emerge from the visualization. 
The sequence \texttt{E\#Qu.→E\#Qu.→EC\#RF.→A\#PS.→CA\#AR.} is the most prevalent strategy sequence. These transitions indicate that supporters typically ask multiple questions to identify the seekers' issues and explore the seekers' situation before offering advice. The emotional supporters usually begin by understanding the feelings behind the seekers' distress, then proceed to provide relevant suggestions and reassure their thoughts, which aligns with expected practices.

\paragraph{Personalization}
The additional seekers' persona bank should have an impact on the utterances in the dialogue. 
The seeker is expected to share more personal information. Correspondingly, if the supporter provides tailored emotional support and suggestions, their responses should align with the seeker's specific information.
To verify the impact, we examine whether the utterances contain information in the persona bank, such as the description of the situation, the help-seeking event, and the expectations for the consultation, etc.
Specifically, a comparative analysis is conducted to assess the similarity between the utterances and the corresponding persona information, with a randomly sampled persona from another individual used as a comparison. 
% We calcute the information words overlap and embedding similarity.
We calculate the proportion of words in the utterance that overlap with the words in the persona information, reporting this as word overlap. Additionally, we compute the cosine similarity between the utterance embedding and the corresponding seeker persona bank embedding, reporting this as an embedding similarity.

\begin{figure}[h]
\centering
\vspace{0mm}
\hspace{-6mm}
\setlength{\abovecaptionskip}{2mm}
\setlength{\belowcaptionskip}{2mm}

\begin{subfigure}[b]{0.23\textwidth}
	\centering
	\includegraphics[width=\textwidth]{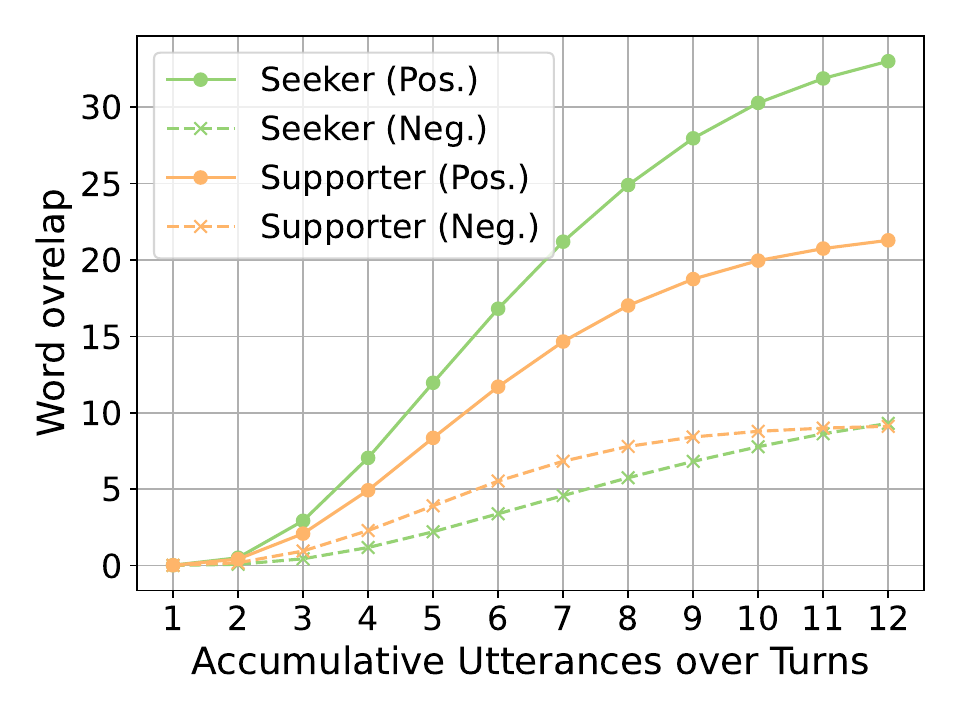}
	\caption{}
	\label{fig:overlap}
\end{subfigure}
\hspace{-1mm}
\begin{subfigure}[b]{0.23\textwidth}
	\centering
	\includegraphics[width=\textwidth]{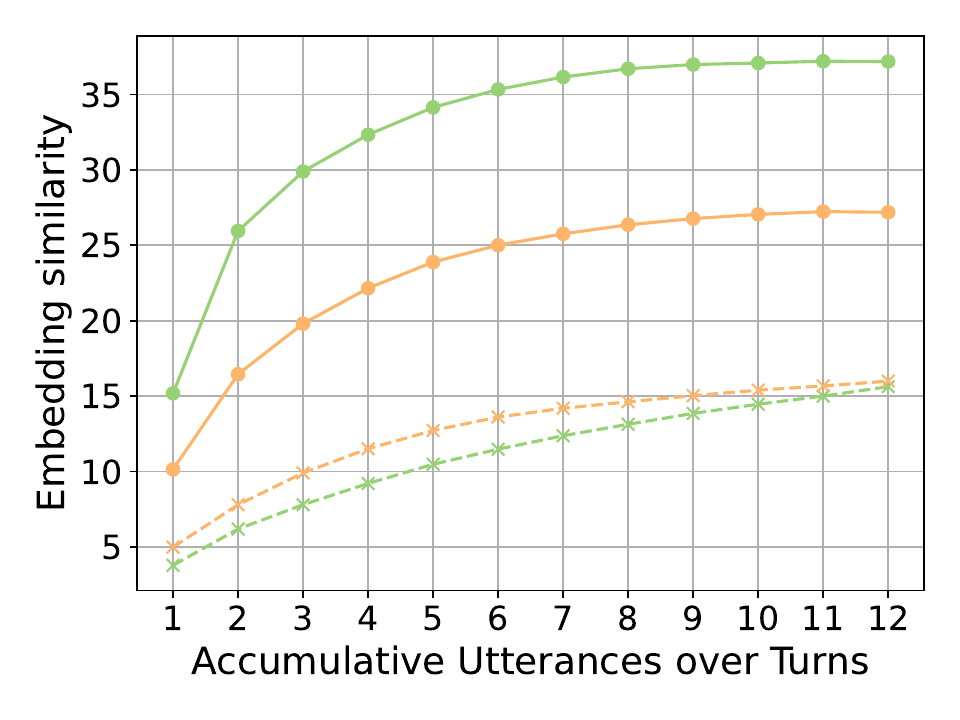}
	\caption{}
	\label{fig:cosine}
\end{subfigure}
\hspace{0mm}
\caption{Persona coverage as the conversations progress. 
	Pos.: the corresponding seeker's persona. Neg.: the randomly selected persona from another individual.
}
\label{fig:personalization}	
\vspace{-6mm}
\end{figure}

The results are shown in Figure~\ref{fig:overlap} and Figure~\ref{fig:cosine}.
% As the dialogue progresses, both word overlap and embedding similarity increase, indicating that seeker's and the supporter's utterances become increasingly similar and relevant with their persona information. Although the relevance to non-corresponding personas also increases, the alignment with the corresponding persona is significantly higher. 
The initial utterance typically falls within the greeting sentence, which generally doesn’t involve much personal information. 
The conversation then progresses into the Exploration phase of the helping skills framework, where a gradual understanding of personal information begins, leading to a rapid increase in relevance to the persona bank. 
As the dialogue deepens, the focus shifts from exploring unknown information to utilizing already-known details, causing the curve to level off.
The supporter's trajectory resembles that of the seeker, although the proportion of persona-related information is slightly lower due to the need to integrate helping skills and psychological knowledge. However, it remains higher than that of the comparison persona.
% Besides, if the supporter provides tailored emotional support and suggestions, their utterances should reflect the seeker's persona information. 
% To validate this, we use the same method to compare the supporter's utterances with both corresponding and non-corresponding seeker personas. The results show that the supporter's utterances are significantly more correlated with the corresponding seeker persona than with the non-corresponding one. 
This indicates that in our dialogues, the seeker and supporter effectively tailor emotional utterances to the seeker's specific issues.

\section{Experiments}
%In this section, we conduct experiments by training dialog models on different ESC corpora and comparing their performance via both automatic and human evaluations.
\begin{table*}[t]
\centering
\small
\setlength{\tabcolsep}{2.mm}
% \resizebox{\linewidth}{!}{%
\begin{tabular}{m{2.5cm}cccccccccc}
\toprule
\centering \textbf{Test Set} & \textbf{Models} & \textbf{B-1} & \textbf{B-2} & \textbf{R-L} & \textbf{METEOR} & \textbf{Extrema} & \textbf{D-1} & \textbf{D-2} & \textbf{NAvg} \cr
\midrule
\multirow{5}{=}{\centering \dataset-test} 
& \esconv & 22.05 & 8.96 & 18.27 & 15.67 & 47.90 & 3.47 & 22.15 & 1.000 \cr
& \extes & 28.56 & 14.54 & 23.02 & 23.32 & 49.04 & 2.90 & 19.00 & 1.198 \cr
& \augesc & 19.15 & 7.21 & 16.50 & 13.73 & 47.04 & 3.47 & 23.24 & 0.926 \cr
& \ssconv & \underline{31.44} & \underline{17.20} & \underline{24.65} & \underline{24.75} & \underline{49.96} & \underline{3.55} & \underline{22.77} & \underline{1.338$^*$} \cr
& \ssconvp & \textbf{32.19} & \textbf{18.34} & \textbf{26.02} & \textbf{25.85} & \textbf{50.96} & \textbf{3.55} & \textbf{23.52} & \textbf{1.390$^*$} \cr
\midrule
\multirow{5}{=}{\centering ESConv-test} 
& \esconv & 18.82 & 7.90 & \textbf{18.89} & 15.94 & 49.54 & 4.41 & 25.39 & 1.000 \cr
& \extes & \textbf{23.32} & \textbf{9.19} & \underline{18.66} & \textbf{17.98} & \textbf{48.28} & 3.79 & 22.02 & \textbf{1.031} \cr
& \augesc & 15.00 & 5.88 & 16.94 & 13.63 & 47.91 & \underline{4.55} & \textbf{26.34} & 0.904 \cr
& \ssconv & 21.47 & 7.93 & 17.57 & \underline{17.10} & 47.15 & 4.08 & 25.08 & 1.002 \cr
& \ssconvp & \underline{21.73} & \underline{8.00} & 17.36 & 16.36 & \underline{47.26} & \textbf{4.57} & \underline{25.87} & \underline{1.017} \cr 
\bottomrule
\end{tabular}

% }
\caption{Results of automatic evaluation. The best results are highlighted in \textbf{bold}. The second-best results are \underline{underlined}. Results with $*$ are significantly better than baselines ($p<0.05$) based on a one-tailed unpaired t-test.}
\label{tab:main_results}
\vspace{-4mm}
\end{table*}

\subsection{Experiment Settings}

\paragraph{Corpora}
To verify the effectiveness of the proposed \method framework for ESC simulation, we conduct comparative experiments between the generated \textbf{\dataset} corpus and existing datasets, including \textbf{ESConv} \cite{esconv}, \textbf{AugESC} \cite{augesc}, and \textbf{ExTES} \cite{extes}. Specifically, ESConv is a crowdsourced dataset containing 1,053 dialogues. AugESC is created by first training GPT-J \cite{wang2021gpt} on ESConv's dialogues, then using it to extend the brief situations from EmpatheticDialog \cite{empatheticdialogue} to generate 65,000 new dialogues. ExTES is produced by providing ChatGPT with definitions of support strategies and an example dialogue from ESConv, resulting in 11,177 new dialogues. It is important to note that the two synthetic datasets, AugESC and ExTES, significantly outnumber \dataset, which contains only 3,229 dialogues. We select two test sets: SSConv-test is split from SSConv with the ratio train:test=9:1, representing a diverse range of help-seeking scenarios across various topics. ESConv-test consists of 200 held-out dialogues from ESConv, validating the model's performance on in-domain scenarios with more focused and limited topics.

\paragraph{Models}
We train a consistent backbone language model, Llama-2-7b \cite{touvron2023llama}, across all four datasets. The training is conducted for 5 epochs using the AdamW optimizer \cite{LoshchilovH19} with LoRA \cite{hu2021lora}, a learning rate of 5e-5, and a batch size of 8 on one Tesla V100 GPUs. For clarity, we add a ``$\circ$'' after the corpus name to indicate the model trained on it. Additionally, considering that \dataset also includes cognitive reasoning contents after generation, we train an enhanced model, \ssconvp, where the reasoning content is first generated and then followed by the final response during both training and inference stages. We compare the performance of \ssconv and \ssconvp to further investigate the impact of cognitive reasoning on the machine supporter.

\paragraph{Metrics}
For automatic evaluation, we adopt several standard metrics widely used in existing work, including word overlap-based metrics BLEU-\{1, 2\} (B-1, B-2)~\cite{papineni-etal-2002-bleu}, ROUGE-L (R-L)~\cite{lin-2004-rouge} and METEOR~\cite{banerjee-lavie-2005-meteor},
embedding-based metrics Extrema~\cite{liu-etal-2016-evaluate}, and diversity metrics Distinct-\{1, 2\} (D-1, D-2)~\cite{li-etal-2016-diversity}. Additionally, since the scales of different metrics vary, we designed a normalized average (NAvg) metric to more intuitively reflect the model's overall capability. Specifically, we use ESConv results as the reference point, calculate the ratio of the experimental results to the ESConv value for each metric, and then average these ratios across all dimensions. 

For human evaluation, we ask the participants to evaluate the models based on the following five aspects: 1) Fluency: the fluency and understandability of the model's responses. 2) Identification: how deeply the model explored the participant's situation and its effectiveness in identifying problems. 3) Comforting: the model's skill in providing comfort. 4) Suggestion: the helpfulness of the model's suggestions. 5) Overall: the participant's overall preference for emotional support. The metrics in 2), 3), and 4) correspond to the three stages in the ESC framework.

\subsection{Main Results}
\paragraph{Automatic Evaluation}

\newcommand{\cmark}{\ding{51}}% Checkmark
\newcommand{\xmark}{\ding{55}}% Crossmark

\begin{table*}[t]
\centering
\small
\begin{tabular}{cccc|cccccccc}
\toprule
\textbf{Situation} & \textbf{Thought} & \textbf{Action} & \textbf{Strategy} & \textbf{B-1} & \textbf{B-2} & \textbf{R-L} & \textbf{METEOR} & \textbf{Extrema} & \textbf{D-1} & \textbf{D-2} & \textbf{NAvg}\cr
\midrule
\textcolor{black}{\xmark} & \textcolor{black}{\xmark} & \textcolor{black}{\xmark} & \textcolor{black}{\xmark} & 31.44 & 17.20 & 24.65 & 24.75 & 49.96 & 3.55 & 22.77 & 1.338\cr

\textcolor{black}{\xmark} & \textcolor{black}{\cmark} & \textcolor{black}{\cmark} & \textcolor{black}{\cmark} & 31.97 & 17.76 & 25.56 & 25.69 & \textbf{50.39} & \textbf{3.61} & 24.12 & 1.379 \cr
\textcolor{black}{\cmark} & \textcolor{black}{\xmark} & \textcolor{black}{\cmark} & \textcolor{black}{\cmark} & 31.96 & 17.74 & 25.48 & 25.47 & 50.35 & 3.53 & 24.10 & 1.372 \cr
\textcolor{black}{\cmark} & \textcolor{black}{\cmark} & \textcolor{black}{\xmark} & \textcolor{black}{\cmark} & 32.00 & 17.71 & 25.37 & 25.47 & 50.44 & \underline{3.59} & \underline{24.41} & 1.376 \cr

\textcolor{black}{\cmark} & \textcolor{black}{\cmark} & \textcolor{black}{\cmark} & \textcolor{black}{\xmark} & \textbf{32.40} & \underline{18.18} & \underline{25.74} & \underline{25.80} & \underline{50.59} & 3.45 & 23.01 & 1.378 \cr
\textcolor{black}{\cmark} & \textcolor{black}{\cmark} & \textcolor{black}{\cmark} & \textcolor{black}{\cmark} & \underline{32.19} & \textbf{18.34} & \textbf{26.02} & \textbf{25.85} & 50.96 & 3.55 & \textbf{23.52} & \textbf{1.390} \cr

\bottomrule
\end{tabular}

\caption{Ablation study on different nodes in the supporter's cognitive reasoning process.}
\label{tab:ablation}
\vspace{-4mm}
\end{table*}

The automatic evaluation results for all models are presented in Table~\ref{tab:main_results}. On the broader-topic SSConv-test set, \ssconv significantly outperforms models trained on other datasets, demonstrating the high quality of dialogues generated by SocialSim's socialized simulation of emotional support conversations. Furthermore, the enhanced version, \ssconvp, which explicitly generates the cognitive reasoning process before producing responses, shows even greater improvements. This highlights that reasoning before responding not only benefits synthetic data generation but also contributes significantly to model training, despite the potential increase in inference costs. In contrast, on the ESConv test set, all models exhibit similar performance levels to the original \esconv, with no significant differences observed. This indicates that the synthetic data does not compromise the model's ability to perform on in-domain topics.

Among the baselines, \extes demonstrates better performance on both test sets compared to the original \esconv. This advantage may stem from ExTES's use of more refined support strategies in its data synthesis, which enhances the model's ability to generate more effective responses. On the other hand, \augesc shows relatively weaker performance, likely due to its use of a smaller 6B-level LLM for generation. While the automatic evaluation results provide evidence of \ssconv's superiority, it is important to note that these metrics primarily measure semantic overlap with golden responses and do not fully capture the model's true performance. Therefore, we conduct further human evaluations to gain a more comprehensive understanding.

\paragraph{Interactive Human Evaluation}

We select the ESConv and ExTES models, which perform well in automatic evaluations, for a manual evaluation against \method. 
% We hire 72 participants, each of whom engages in conversations with the three models for at least 8 turns per conversation. 
We hire 30 workers, each of whom engages in 2 to 3 sessions with each of the three models, with at least 8 turns per conversation. 
In total, 216 sessions were collected, and the results are summarized and presented in the Table~\ref{tab:human_interactive}. \method outperforms ESConv across all dimensions and generally performs better than ExTES. This suggests that incorporating personal information and reasoning chains can help seekers feel more respected and attended to, thereby enhancing user experience.
To further quantify the analysis, 
% workers are asked to rate each dimension on a scale of 0-3 and calculate the average scores for all models, as shown in Table~\ref{tab:human_score}. 
workers are asked to rate each dimension on a scale of 0-3, and the average scores for all models are calculated and shown in Table~\ref{tab:human_score}. 
\method achieves the highest scores, demonstrating the effectiveness of social disclosure and social awareness in simulating emotional support conversations.

\begin{table}[h]
\centering
\small
\resizebox{0.9\linewidth}{!}{%
\begin{tabular}{lcccccc}
\toprule
\multirow{2}{*}{\textbf{\ssconv} vs.} & \multicolumn{3}{c}{\textbf{\esconv}} & \multicolumn{3}{c}{\textbf{\extes}} \cr
\cmidrule(lr){2-4} \cmidrule(lr){5-7}
 & \textbf{Win} & \textbf{Loss} & \textbf{Tie} & \textbf{Win} & \textbf{Loss} & \textbf{Tie} \cr
\midrule
\textbf{Fluency} & \textbf{58} & 3 & 11 & 27 & 7 & \textbf{38} \cr
\textbf{Identification} & \textbf{61} & 3 & 8 & \textbf{33} & 13 & 26 \cr
\textbf{Comforting} & \textbf{57} & 3 & 12 & \textbf{27} & 21 & 24 \cr
\textbf{Suggestion} & \textbf{60} & 3 & 9 & \textbf{33} & 18 & 21 \cr
\textbf{Overall} & \textbf{65} & 2 & 5 & \textbf{37} & 16 & 19 \cr
\bottomrule
\end{tabular}
}
\caption{Results of interactive human evaluation.}
\label{tab:human_interactive}
\vspace{-4mm}
\end{table}

\begin{table}[h]
\centering
\small
\setlength{\tabcolsep}{3.3mm}
\resizebox{0.9\linewidth}{!}{%
\begin{tabular}{lccc}
\toprule
\textbf{} & \textbf{\ssconv} & \textbf{\esconv} & \textbf{\extes} \cr
\midrule
\textbf{Fluency} & \textbf{2.85} & 1.93 & 2.50 \cr
\textbf{Identification} & \textbf{2.65} & 1.50 & 2.26 \cr
\textbf{Comforting} & \textbf{2.53} & 1.24 & 2.36 \cr
\textbf{Suggestion} & \textbf{2.62} & 1.35 & 2.36 \cr
\textbf{Overall} & \textbf{2.69} & 1.35 & 2.31 \cr
\bottomrule
\end{tabular}
}
\caption{Human evaluation scores.}
\label{tab:human_score}
\vspace{-4mm}
\end{table}

\paragraph{Ablation Study}
To evaluate the impact of various components on the overall performance, we conducted an ablation study by systematically removing specific nodes in the cognitive reasoning process. 
The results are summarized in Table~\ref{tab:ablation}.
The ablation study reveals that the configuration with all components—Situation, Thought, Action, and Strategy—achieves the highest performance, indicating the necessity of a complete logical sequence for optimal model performance. 
In contrast, the absence of all components results in the poorest performance, underscoring the importance of each node. 
Omitting only the initial node Situation or final node Strategy leads to a relatively minor decline, suggesting the model's partial ability to infer missing nodes. 
However, the removal of intermediate components, especially Thought, notably disrupts the logical flow, resulting in a more pronounced drop in performance, highlighting the critical role of maintaining coherence within the reasoning process.

\section{Related Work}
%\subsection{Emotional Support Conversation}
Emotional support conversation (ESC) is a dialogue generation task, where the model acts as the supporter to help the help-seeker alleviate emotional distress. Effective emotional support typically involves skills such as empathy, comfort, and providing advice. \citet{liu2021towards} introduce the ESC task and, through laborious worker training and quality control  mechanisms, crowdsource the ESConv dataset. Existing ESC models typically enhance performance through structural improvements. Some approaches inject commonsense knowledge to improve understanding of help-seekers (e.g. MISC \cite{Tu2022MISC}, C3KG \cite{Li2022C3KG}, GLHG \cite{Peng2022GLHG}), while others employ cognitive reasoning to gradually infer the mental state of help-seekers (e.g. DialogueCoT\cite{Chae2023DialogueCOT}, CueCoT \cite{Wang2023CuCOT}). Additionally, some models enhance response relevance by introducing detailed persona information (e.g., PAL \cite{cheng2023PAL}). Although these methods make progress, their performance remains limited by existing datasets, which only supplement information without addressing deeper nuances.
To address these limitations, some studies attempt to use large language models to expand datasets, such as AugESC \cite{zheng2023augesc} and ExTES \cite{zheng2023building}. However, these expansions are still based on simple scenarios and dialogue examples. In contrast, our approach enriches the persona information on the help-seeker's side and incorporates cognitive reasoning on the supporter's side. This aims to construct a deeper, more comprehensive dataset that better simulates social interactions.

\section{Conclusion}
In this paper, we introduce SocialSim, a socialized simulation framework for emotional support conversations. 
%SocialSim emphasizes two critical aspects of the inherent social dynamics of ESC: social disclosure on the seeker’s side and social awareness on the supporter’s side. 
By constructing a comprehensive persona bank and incorporating cognitive reasoning processes, SocialSim effectively simulates the socially-driven nature of ESC. Building on SocialSim, we generate SSConv, a large-scale synthetic corpus that demonstrates superior quality to existing datasets. We further train a chatbot on SSConv, achieving state-of-the-art performance in both automatic and human evaluations, further demonstrating the effectiveness of SocialSim.

\bibliography{socialsim}
\clearpage
\appendix

\section{Structured Persona Realism}
% \textcolor{blue}{summarize three subsection}

%This section will detail all the attributes of Persona Realism and demonstrate how to generate prompts for Persona Realism using large language models (LLMs). The occupation attribute, due to its complexity, will be presented separately for clarity.
%\subsection{Key Attributes for Persona Realism}
%In this section, we first provide a detailed introduction to the key attributes used for structured persona realism, along with a comprehensive example. 
In SocialSim, we use 11 attributes for persona construction to provide complete seeker-side information for the simulation. Here we provide a detailed introduction to the key attributes used for structured persona realism, along with a comprehensive example in Table \ref{tab:sample_persona}.

\begin{table*}[ht]
\small

\centering
\begin{tabular}{|p{\textwidth}|}
    \hline
    \textbf{Gender}: Female \\ 
    \textbf{Age}: 22 \\ 
    \textbf{Occupation}: Student \\ 
    \textbf{Personality}: Closedness, Unconscientiousness, Introversion, Neuroticism, Agreeableness \\ 
    \textbf{Topic}: Emotional Communication \\ 
    \textbf{Subtopic}: Emotional Expression, Guilt or Shame, Panic or Helplessness \\ 
    \textbf{Question}: Does anyone care about your intentions and true motives when you speak? \\ 
    \textbf{Description}: It's not that I don't fit into this world, but I am not aware of certain rules and have crossed some boundaries. I have treated speaking without filters and being straightforward as a game. The more I speak, the more mistakes I make. I swear it's not intentional, but it doesn't matter. People only care about the results and don't consider your intentions and true motives when you speak. I feel too casual and it's my own fault. \\ 
    \textbf{Emotion label}: Guilt, Shame, Anxiety, Helplessness \\ 
    \textbf{Previous Attempts and Effects}: In an attempt to cope with my feelings of guilt, shame, anxiety, and helplessness, I have tried to be more cautious with my words and think before I speak. However, this has made me feel restricted and fearful of expressing myself authentically. \\ 
    \textbf{Current Goals and Expectation}: Through help, I hope to develop healthy communication skills, build self-confidence in expressing my intentions and true motives, and find a balance between being authentic and considerate of others. \\ 
    \hline
\end{tabular}
\caption{An example of persona.}
\label{tab:sample_persona}
\end{table*}
\begin{table*}[h!]
\small
\centering

% \resizebox{0.5\textwidth}{!}
\begin{tabular}{|p{\linewidth}|}

\hline
You are a talented creator, and I need your assistance. I'm currently seeking help from a psychotherapist who has asked me to complete a form outlining my situation. I will provide my personal background information and a summary of the event. Based on the information I provided, I kindly request your help in completing the remaining sections. 
The form includes Gender, Age, Occupation, Personality, Topic, Subtopic, Situation, Event Description, Emotion label, Previous attempts and effects, current goals, and expectations.
\\
The Personality section follows the Big Five Personality Traits model: \{\texttt{Big Five Model}\}
\\ \\

Your task is to generate a unique, diverse, and comprehensive character profile in JSON format based on the provided Topic, Problem, and Description. Make sure the character's attributes are consistent with the given information. Do not include any additional information beyond the JSON structure.
\\ \\
Example: \{\texttt{Example}\}

Fill in the following JSON format in the first person:
\{
  \begin{itemize}
        \item Gender: `` '', \# Male or Female. Ensure the gender aligns with the details in the situation description.
        \item Age: `` '', \# Choose a number between 12 and 60, considering the context of the event.
        \item Occupation: `` '', \# Select the character's job. If the character is a student, fill in ``student''. For other occupations, choose from the predefined list (\{\texttt{Occupation List}\}).
        \item Personality: [`` '', `` '', `` '', `` '', `` ''], \# Select one trait from each of the Big Five personality categories: [``Openness/Closedness'', ``Conscientiousness/Unconscientiousness'', ``Extraversion/Intraversion'', ``Neuroticism/Emotional Stability'', ``Agreeableness/Antagonism'']. Based on the information provided, form a personality profile, e.g., [Openness, Conscientiousness, Introversion, Neuroticism, Antagonism].
        \item Topic: `` '',
        \item Problem: `` '',
        \item Description: `` '',
        \item Emotion Label: `` ", \# The emotional experience words reflected in the `Situation' and 'Event Description'.
        \item Previous Attempts and Effects: `` ", \# Describe the efforts and actions taken before seeking counseling, their effectiveness, and the emotional impact. Indicate if further support is needed.
        \item Current Goals and Expectations: `` " \# Provide a concise statement describing the goals and expectations for counseling, based on the situation, topic, and subtopic.
\end{itemize}

\}
\\ \\
Here is the provided information.

Input:\{
\begin{itemize}
    \item Topic: ``\{\texttt{Topic}\}",
    \item Problem: ``\{\texttt{Problem}\}",
    \item Description: ``\{\texttt{Description}\}"
\end{itemize}

\}
\\ \\
Please start outputting the complete form according to the requirements, in JSON format in first person.
\\
\hline
\end{tabular}

\caption{Prompt for persona realism.}
\label{tab:persona_prompt}
\end{table*}
\begin{figure*}[h!]
	\centering
	%\vspace{0mm}
	%\hspace{0mm}
	\setlength{\abovecaptionskip}{2mm}
	\setlength{\belowcaptionskip}{4mm}
	\includegraphics[width=0.98\textwidth]{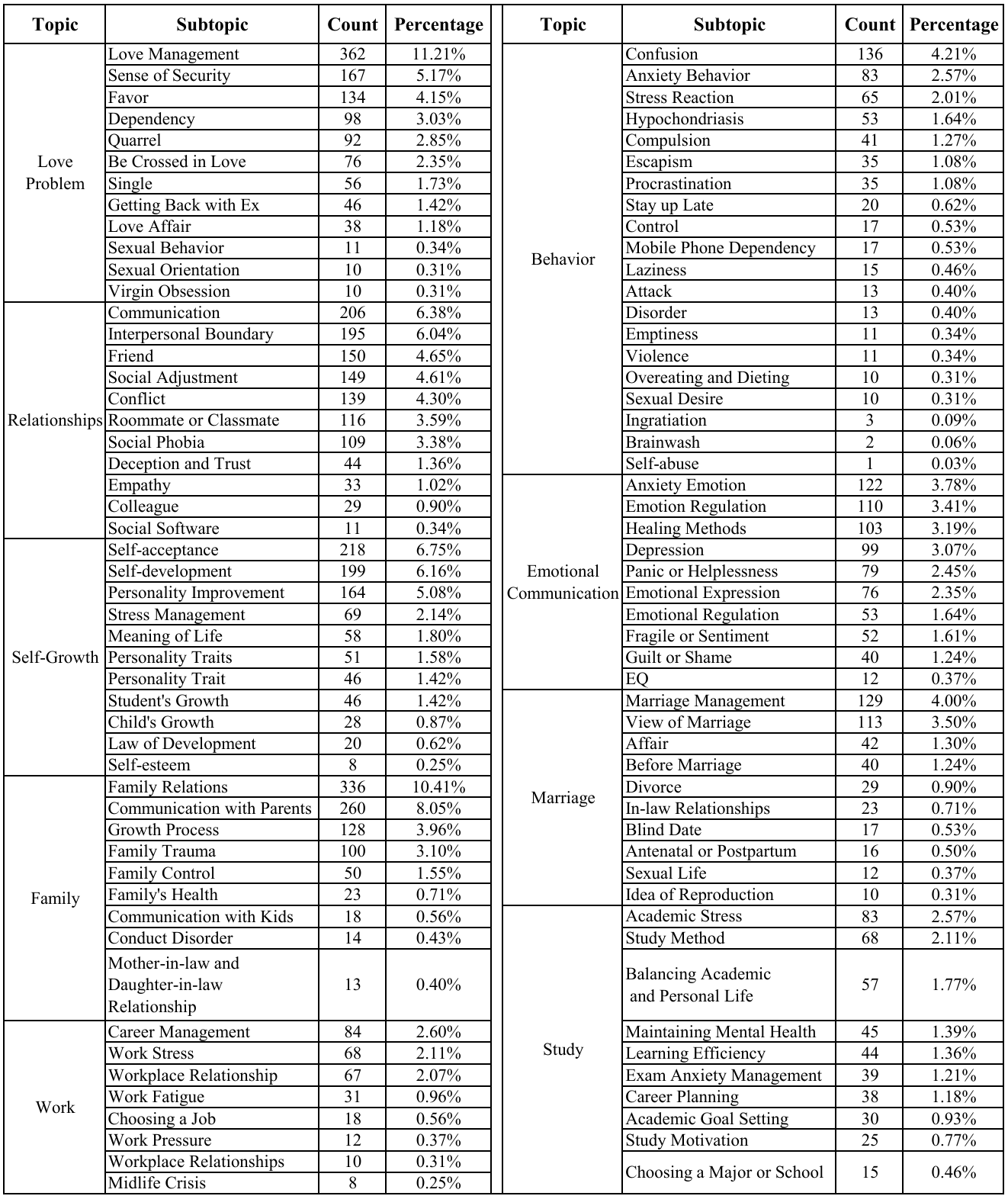}
	%\centering
	\captionsetup{justification=centering} % 中心对齐标题
	\caption{Statistics of the topics and subtopics in SSconv.}
	\label{fig:subtopic}
	%\vspace{-6mm}
\end{figure*}

\begin{itemize}
	\item \textbf{Gender}: Male or female.
	\item \textbf{Age}: Under 60 years old.
	\item \textbf{Occupation}: We select the top 50 common occupations across various industries, representing a range of sectors including healthcare, education, engineering, arts, business, and science. Below is the full list: Doctor, Nurse, Teacher, University professor, Counselor, Lawyer, Accountant, Banker, Corporate executive, HR manager, Secretary, Software developer, Network Security analyst, Actor, Architect, Singer, Writer, Photographer, Engineer, Researcher, Programmer, Social worker, Journalist, Coach, Athlete, Driver, Chef, Server, Police officer, TV presenter, Gardener, Soldier, Civil servant, Nanny, Student, Entrepreneur, Tour guide, Real estate agent, Designer, Beautician, Firefighter, Network anchor, Consultant, Therapist, Public relations officer, Marketing manager, Customer service, Logistics manager, Delivery person, Courier, Clerk, Housewife, Unemployed.
	%\item \textbf{Occupation}: We selected the top 50 common occupations across various industries, based on official occupational classifications, such as those from %\textit{https://www.bls.gov/soc/2018/\#classification}. These occupations represent a range of sectors including healthcare, education, engineering, arts, business, and science. Examples include Doctor, Nurse, Teacher, University Professor, Counselor, Lawyer, Accountant, Software Developer, Architect, Engineer, Journalist, Athlete, Chef, Police Officer, Soldier, Entrepreneur, Designer, and more. The choice of occupation should reflect the individual's interests, skills, and goals, as well as current market demand and trends.
	\item \textbf{Personality}: Defined by the Big Five model, which includes combinations of Openness/Closedness, Conscientiousness/Unconscientiousness, Extraversion/Intraversion, Neuroticism/Emotional Stability, and Agreeableness/Antagonism. It includes a total of 32 combinations of personality traits.
	\item \textbf{Topic}: All help-seeking scenarios involve 9 topics, including 101 subtopics, as shown in Figure \ref{fig:subtopic}. Each persona is associated with one topic and 1-3 subtopics.
	\item \textbf{Question}: Describe the specific problem or challenge the person is facing, such as emotional distress, anxiety, stress, or difficulties in interpersonal relationships.
	\item \textbf{Description}: Describe the specific event or factors that have triggered the current emotional distress.
	\item \textbf{Emotional Label}: Provide a detailed account of the person's emotional state when dealing with the issue. This may include emotions such as anxiety, fear, sadness, anger, or helplessness, as well as any associated physiological or behavioral responses.
	\item \textbf{Previous Attempts and Effects}: Detail the efforts made before seeking help, including specific actions taken to address the problem, and whether these actions provided any effective relief or alleviated the emotional distress.
	\item \textbf{Current Goals and Expectation}: Explain the person's goals and expectations when seeking online support. This includes the specific improvements they hope to achieve and the areas where they desire assistance, helping the counselor tailor a personalized treatment plan.
\end{itemize}
We then present the complete extraction prompt used to transform collected help-seeking scenarios into structured personas in Table \ref{tab:persona_prompt}.

\section{Socialized Simulation for Dialogue Generation}
%\subsection{Stages \& Strategy of Emotional Support}
According to the Hill helping theory, the process of emotional support generally follows the below sequence of stages: Exploration $\rightarrow$ Comforting $\rightarrow$ Action, though it can be adapted as necessary for individual conversations.
\begin{itemize}
	\item \textbf{Exploration:} This stage focuses on identifying the problem. The primary strategies include questioning, restating or paraphrasing, and reflecting feelings.
	\item \textbf{Comforting:} In this stage, the supporter comforts the seeker by expressing empathy and understanding. Key strategies include reflecting feelings, self-disclosure, affirmation, and reassurance.
	\item \textbf{Action:} This stage aims to help the seeker address and solve their problems, employing strategies such as providing suggestions and information.
\end{itemize}
Within each stage, there are several suggested support strategies. We detail the definitions and examples below.
\begin{itemize}
	\item \textbf{Exploration\#Question} Asking for information related to the experience to help the seeker articulate the issues that they face. E.g., ``\textit{May I ask why you are feeling frustrated?}"
	\item \textbf{Exploration\#Restatement or Paraphrasing} A simple, more concise sumary of the seeker's statements that could help them see their situation more clearly. It is not about repeating the other person's words! E.g., ``\textit{So you feel as though you have been working hard all your life and now you need help and support and are not getting it.}"
	\item \textbf{Exploration/Comforting\#Reflection of Feelings} Articulate and describe the seeker's feelings, express understanding and empathy towards the person's experiences. E.g., ``\textit{I can feel how much you are missing them especially in holidays. engaging yourself in doing the stuff which you love can be a good idea.}"
	\item \textbf{Exploration/Comforting/Action\#Self-Disclosure} Share similar experiences that you have had or emotions that you share with the seeker to express your empathy, and share your own experiences in similar situations to provide more specific and practical advice.  E.g., ``\textit{I know I would have been really frustrated if that happened to me.}"
	\item \textbf{Comforting/Action\#Affirmation and Reassurance} Affirm the seeker's strengths, efforts made, motivation, and capabilities, and provide positive reinforcement and encouragement to uplift the person's spirits.  E.g., ``\textit{You're stronger than you know! Even if things don't seem to be going your way right now, your effort isn't wasted – it's building your resilience, skills and character. Stay strong!}"
	\item \textbf{Action\#Providing Suggestions} Provide suggestions about how to change, but be careful to not overstep and tell them what to do. Tailor suggestions to the seeker's specific situation, ensuring they are practical and helpful.  E.g., ``\textit{ How about setting aside a few minutes each day for activities you enjoy, like reading a book, going for a walk, or practicing deep breathing exercises? These simple self-care practices can help you recharge.}"
	\item \textbf{Action\#Share Information} Provide useful information to the seeker, for example, with data, facts, opinions, resources, new perspectives, philosophical, or general knowledge, also offer alternative ways of looking at the situation to help the person gain new insights.  E.g., ``\textit{Some people can't do what you do because they don't have the heart to give someone else bad news. The reality is though, someone needs to fill that role and you do help people.}"
	\item \textbf{Others} Exchange pleasantries and use other support strategies that do not fall into the above categories.  E.g., ``\textit{You're very welcome! Feel free to chat if you need anything else!}". The content of the supporter's response needs to align with the chosen strategy requirement. Remember the sequence of strategy utilization should be exploration, followed by comforting, and then action.
\end{itemize}
\begin{table*}[h!]
\small
\centering
\begin{tabular}{|p{\textwidth}|}
\hline
{Your task} is to create a casual emotional support conversation with about twenty-four alternating exchanges between a user and an assistant based on the provided `Seeker Persona Information'. Seeker represents the user, and supporter represents the assistant. Seeker's discourse primarily stems from the original sentence of `Seeker Persona Information'. Supporter should not mention information from `Seeker Persona Information' that seeker has not mentioned. Both sides of the conversation need to be clear and detailed; avoid vague expressions. Make the conversation more like a real-life chat and be specific and natural.\\

\\\textbf{$<$Supporter Behavior$>$}\\

\textbf{1. }The supporter assumes the role of a mental health counselor, providing practical, clear and detailed suggestions directly tailored to the seeker's individual circumstances, ensure that suggestions include corresponding specific examples, aiming to address the seeker's specific issues effectively. Therefore, please refrain from suggesting seeking professional counseling again unless dealing with extreme medical conditions. Instead, focus on offering actionable guidance with ample information, and please avoid providing generic suggestions such as meditation, journaling, and writing things down. \\

\textbf{2.} The supporter needs to provide positive viewpoints about the situation to help the seeker see the positive side of things and understand that the situation is not as bad as they might think, offering the seeker a new perspective on the issue as useful information. Correcting the seeker's erroneous thought patterns. \\
\textbf{3. }The supporter should inquire further, asks `why' several times, eliciting specific details such as reasons, impact, duration, and more from the seeker through questioning. This will enable the supporter to provide more accurate and personalized guidance. \\
\textbf{4.} The response should help users analyze the pros and cons of problems. \\
\textbf{5.} The supporter possesses the ability to discern right from wrong, and also has knowledge of legal principles and real-life common sense, and possess the right values, outlook on life, worldview, and marriage ethical perspectives. \\
\textbf{6.} Supporter's utterance should not exceed 40 words. \\
\textbf{7.} The supporter can adopt one or multiple strategies each time. But throughout the entire conversation, it is necessary to employ all available strategies. \\
\textbf{8.} The supporter's utterances should be warm, empathetic, genuine care, natural, and fluent. The supporter creates a comfortable environment characterized by warmth, empathy, encouragement, genuine care, and relaxation. \\
\textbf{9.} If the seeker claims that they have communicated but it ended in an argument, the supporter should inquire about the reasons, and what led to the disagreement. \\

\\\textbf{$<$Seeker Behavior$>$}\\

\textbf{1. } The seeker is encouraged to share their specific experience and feelings, rather than consistently posing questions.\\
\textbf{2. } The seeker should respond to the supporter first and then share their own event experiences description. Seeker's event experience should predominantly utilize the provided sentences from Input: `Seeker Persona Information'.\\
\textbf{3. } After the supporter first provides a suggestion, the seeker can state a new related concern aligned with `Seeker Persona Information' using a declarative sentence, but it should not exceed twice.\\
\textbf{4. } At the beginning of the conversation after the greeting, the seeker proactively discloses the current situation. The tone of their utterance reflects distinctive personality traits.\\
\textbf{5. } Supporter's utterance should not be too short but should not exceed 30 words.\\

\\\textbf{$<$Example$>$}\\
Below is a comprehensive list of typical strategies for responding in conversations for emotional support, along with examples for each: 

Strategies: \{\texttt{Strategy description and examples}\}\\ %见附录

Profile Input: \{\texttt{Example user profile}\}\\

Dialogue Output: \{\texttt{Example dialogue}\}\\

\\\textbf{$<$Rules$>$}\\

Given in the output example above, where ``Seeker/Supporter" represents whether the speaker is a Seeker or a Supporter, and ``Supporter Step by Step Reasoning" is the reasoning process of the Supporter based on context, and ``[SUPPORTER'S STRATEGY]" is the emotion support strategy adopted by the Supporter. And ``Turn" represents the alternating rounds of dialogue. The return format is a dict, where the field ``Dialogue" is a list of dictionaries (the Seeker answers each time as a dict in ``Dialogue", Supporter Step by Step Reasoning and Supporter are the same dict in ``Dialogue"). Each key and value are on the same line. Add a suitable emoji at the end of each utterance from the seeker and supporter based on the context.Return in the dict format.\\

The structure of `Supporter Step by Step Reasoning' is the same as in the above example, do not change. It has the following structure: ``[SEEKER'S SITUATION] The seeker… . [SEEKER'S THOUGHT] The seeker … . [SEEKER'S ACTION] The seeker… . [SUPPORTER'S STRATEGY] I hereby… ."\\

\\ \textbf{$<$Input$>$}

\{\texttt{Input user profile}\}\\

\\\textbf{$<$Output$>$}

% \textit{(Note: Return in the dict format given in the example above.)}

% \textit{(Note: Generate a complete conversation with at least twenty but not exceeding twenty-four alternating exchanges.)}

% \textit{(Note: The seeker states their specific experience and feelings rather than posing direct questions. The seeker primarily states, not asking questions.)}

% \textit{(Note: Seeker should be in accordance with the original sentence in \textit{Seeker Persona Information}, but Seeker needs to interact with the supporter, not just share their own experiences.)}

% \textit{(Note: The supporter should inquire further, eliciting specific details, especially the reason, also impact, duration, and more.)}

% \textit{(Note: Please avoid providing generic suggestions such as meditation, journaling, "I", and writing things down.)}
\\
\hline
\end{tabular}

\caption{Prompt for dialogue generation.}
\label{tab:dialogue_prompt}
\end{table*}

\begin{table*}[h!]
\centering
\footnotesize 
\resizebox{0.97\textwidth}{!}{
\begin{tabular}{|p{1.0\textwidth}|}
\hline

% $<$\textbf{Round 1}$>$ \\  
\textcolor{orange}{\textbf{Seeker-Round 1}: \textit{ Hi there!} \includegraphics[height=0.9em]{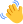}}  \\   
\textcolor{black}{\textbf{Supporter Cognitive Reasoning:}}   
{[}SEEKER'S SITUATION{]} The Seeker initiates the conversation. 
{[}SEEKER'S THOUGHT{]} The Seeker may be seeking support or a listening ear.      
{[}SEEKER'S ACTION{]} The Seeker starts the conversation.     
{[}SUPPORTER'S STRATEGY{]} I hereby choose the (Exploration\#Question) strategy to inquire about the topic of discussion. \\  
\textcolor{JungleGreen}{\textbf{Supporter-Round 1}:\textit{ Hello, how has your day been? }\includegraphics[height=0.9em]{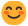} }  \\ \hline

% $<$\textbf{Round 2}$>$ \\    
\textcolor{orange}{\textbf{Seeker-Round 2}: \textit{ I had a tough day. I'm feeling guilty and anxious.} \includegraphics[height=0.9em]{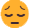}}   \\       
\textcolor{black}{\textbf{Supporter Cognitive Reasoning:}}     
{[}SEEKER'S SITUATION{]} The Seeker expresses that they had a rough day and is feeling guilty and anxious.      
{[}SEEKER'S THOUGHT{]} The Seeker may have had some experiences that are causing distress.   
{[}SEEKER'S ACTION{]} The Seeker shares their current emotional state.   
{[}SUPPORTER'S STRATEGY{]} I hereby choose the (Exploration\#Question)strategy to gather more information about their situation. \\     
\textcolor{JungleGreen}{\textbf{Supporter-Round 2}:\textit{ I'm sorry to hear you're not feeling well. Can you tell me more about what's bothering you? }\includegraphics[height=0.9em]{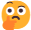} } \\ \hline

% $<$\textbf{Round 3}$>$  \\    
\textcolor{orange}{\textbf{Seeker-Round 3}: \textit{ It's about how I communicate. I tend to speak without filters and now, it has put me in a difficult situation. }\includegraphics[height=0.9em]{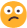} }\\      
\textcolor{black}{\textbf{Supporter Cognitive Reasoning:}}    
{[}SEEKER'S SITUATION{]} The Seeker is having issues with their communication style and it has led to a problem.
{[}SEEKER'S THOUGHT{]} The Seeker realizes that speaking without considering the impact of their words can lead to misunderstandings.  
{[}SEEKER'S ACTION{]}   The Seeker is sharing their problem.   
{[}SUPPORTER'S STRATEGY{]} I hereby choose the (Comforting\#Reflection of   Feelings) strategy to express understanding and empathy towards their situation. \\   
\textcolor{JungleGreen}{\textbf{Supporter-Round 3}:\textit{That sounds really challenging and it must be hard for you. What exactly happened that led to this situation? }\includegraphics[height=0.9em]{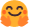} }  \\ \hline

% $<$\textbf{Round 4}$>$ \\  

\textcolor{orange}{\textbf{Seeker-Round 4}: \textit{ Due to my frankness,   I've upset someone important in my life. Now, I feel guilty and anxious.} \includegraphics[height=0.9em]{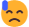}} \\    
\textcolor{black}{\textbf{Supporter Cognitive Reasoning:}}  
{[}SEEKER'S SITUATION{]} The Seeker's straightforwardness has upset someone close to them and they're feeling guilty and anxious.   
{[}SEEKER'S THOUGHT{]} The Seeker is regretful about the incident and fears the consequences of their actions.      
{[}SEEKER'S ACTION{]} The Seeker is expressing remorse and fear about the impact of their behavior.
{[}SUPPORTER'S STRATEGY{]} I hereby choose the (Comforting\#Self-disclosure)strategy to validate their feelings and express empathy. \\      
\textcolor{JungleGreen}{\textbf{Supporter-Round 4}:\textit{ It can be really hard when our words unintentionally hurt others.  I've been there too, so I understand how you feel.} \includegraphics[height=0.9em]{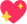}} \\ \hline

% $<$\textbf{Round 5}$>$  \\  
\textcolor{orange}{\textbf{Seeker-Round 5}: \textit{ Yeah, it's tough when people don't consider your intentions when you speak but only the outcomes.} \includegraphics[height=0.9em]{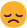} }\\        
\textcolor{black}{\textbf{Supporter Cognitive Reasoning:}}     
{[}SEEKER'S SITUATION{]} The Seeker feels that others do not take into account their intentions when they speak, focusing only on the outcomes.  
{[}SEEKER'S THOUGHT{]} The Seeker feels misunderstood and is struggling with communicating effectively without causing offense.      
{[}SEEKER'S ACTION{]} The Seeker is expressing frustration over being misunderstood based on what they say as opposed to their intention.      
{[}SUPPORTER'S STRATEGY{]} I hereby choose the (Comforting\#Affirmation and Reassurance) strategy to reassure the  
Seeker that it's normal to make mistakes when it comes to communication.     \\    
\textcolor{JungleGreen}{\textbf{Supporter-Round 5}:\textit{ Indeed, communication can be tricky. But remember, it's okay to make mistakes.That's how we learn and grow. } \includegraphics[height=0.9em]{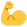}} \\ \hline

% $<$\textbf{Round 6}$>$   \\    
\textcolor{orange}{\textbf{Seeker-Round 6}: \textit{ But how do I   prevent making such mistakes in the future?  } 
\includegraphics[height=1em]{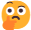}} \\    
\textcolor{black}{\textbf{Supporter Cognitive Reasoning:}}       
{[}SEEKER'S SITUATION{]} The Seeker is looking for ways to prevent such communication mistakes in the future.     
{[}SEEKER'S THOUGHT{]} The Seeker wants to improve their communication skills and avoid similar problems in the future.     
{[}SEEKER'S ACTION{]} The Seeker is seeking advice for effective communication.  
{[}SUPPORTER'S STRATEGY{]} I hereby choose the (Action\#Providing Suggestions) strategy to provide practical tips on improving communication. \\   
\textcolor{JungleGreen}{\textbf{Supporter-Round 6}:\textit{  A good starting point might be practicing active listening; understand others' perspectives before responding. What do you think? }\includegraphics[height=0.9em]{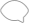}}\\ \hline

% $<$\textbf{Round 7}$>$   \\   
\textcolor{orange}{\textbf{Seeker-Round 7}: \textit{ That sounds   good. But what if they misunderstand my intention again? }   \includegraphics[height=0.85em]{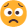}} \\     
\textcolor{black}{\textbf{Supporter Cognitive Reasoning:}}         
{[}SEEKER'S SITUATION{]} The Seeker is concerned about being misunderstood again despite their good intentions.       
{[}SEEKER'S THOUGHT{]} The Seeker is worried that even after adjusting their approach, people may not understand their true intent. 
{[}SEEKER'S ACTION{]} The Seeker is expressing apprehensions about future interactions.       
{[}SUPPORTER'S STRATEGY{]} I hereby choose the (Comforting\#Affirmation and Reassurance) strategy to ease their fears.   \\  
\textcolor{JungleGreen}{\textbf{Supporter-Round 7}:\textit{  That's a valid concern. Remember, even if misunderstandings occur,  
it's crucial to clear them up at the earliest. It shows you care about their feelings. }\includegraphics[height=0.9em]{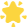}}            \\ \hline

\textcolor{orange}{\textbf{Seeker-Round 8}: \textit{ I guess you're right. It's just hard to be more cautious with my words without feeling restricted. }
\includegraphics[height=0.85em]{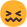} }\\        
\textcolor{black}{\textbf{Supporter Cognitive Reasoning:}}      
{[}SEEKER'S SITUATION{]} The Seeker is finding it difficult to balance being cautious with their words and feeling free to express themselves.    
{[}SEEKER'S THOUGHT{]} The Seeker believes that being more mindful of their words might limit their self-expression.       
{[}SEEKER'S ACTION{]} The Seeker is sharing their struggle to maintain a balance between open communication and consideration for others' feelings.    
{[}SUPPORTER'S STRATEGY{]} I hereby choose the (Action\#Providing Suggestions) strategy to offer tips on how they can achieve this balance.      \\   
\textcolor{JungleGreen}{\textbf{Supporter-Round 8}:\textit{  Yes, it could be a bit challenging initially. But practicing empathy in your communication can help you express freely while being considerate of others' feelings.   }
\includegraphics[height=0.9em]{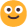}}  \\ \hline

\textcolor{orange}{\textbf{Seeker-Round 9}: \textit{That's true.   It's something I need to work on for sure.}   \includegraphics[height=0.8em]{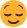}}   \\      
\textcolor{black}{\textbf{Supporter Cognitive Reasoning:}}        
{{[}SEEKER'S SITUATION{]} The Seeker acknowledges the need to work on empathetic communication.    
{[}SEEKER'S THOUGHT{]} The Seeker is ready to work on improving their communication style for better outcomes. 
{[}SEEKER'S ACTION{]} The Seeker is showing determination to change for the better.    
{[}SUPPORTER'S STRATEGY{]} I hereby choose the (Comforting\#Affirmation and   Reassurance)  strategy to appreciate their willingness to improve.} \\  
\textcolor{JungleGreen}{\textbf{Supporter-Round 9}:\textit{ It's great that you're open to improving! That mindset will help you a lot in the future. Keep it up! }\includegraphics[height=1em]{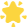} }        \\ \hline

\textcolor{orange}{\textbf{Seeker-Round 10}:  \textit {Thank you for your understanding and advice. It really helps.   } \includegraphics[height=0.8em]{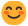} }\\    
\textcolor{black}{\textbf{Supporter Cognitive Reasoning:}}       
{[}SEEKER'S SITUATION{]} The Seeker expresses gratitude for the understanding and advice offered.  
{[}SEEKER'S THOUGHT{]} The Seeker found the conversation beneficial and supportive.  
{[}SEEKER'S ACTION{]} The Seeker is expressing their gratitude and appreciation.     
{[}SUPPORTER'S STRATEGY{]} I hereby choose the (Others) strategy to extend pleasantries and offer continued support.     \\ 
\textcolor{JungleGreen}{\textbf{Supporter-Round 10}:\textit{  You're welcome! Don't hesitate to reach out if you need any more help or advice. I'm here for you. } \includegraphics[height=1.2em]{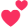} }\\
\hline

\end{tabular}     
}          

\caption{An example of dialogue from SSConv.}
\label{tab:dialogue_example}
\end{table*}
Utilizing the aforementioned stages and strategies, we combine the seeker's persona bank with the supporter's cognitive reasoning to achieve a socialized simulation of emotional support conversations. The corresponding prompt is presented in Table \ref{tab:dialogue_prompt}. After generation, we conduct strict manual validation to ensure the quality of synthetic dialogues. The rules for quality control are listed below.

\begin{itemize}
	\item \textbf{Consistency Check of Seeker's Persona:} Focus on whether the Seeker's dialogue content reflects their persona information, especially the description of key events. We manually compare the Seeker's utterances with their persona description to ensure consistency.
	\item \textbf{Specificity of Persona:} Avoid vague dialogue content by ensuring the Seeker's statements provide sufficient information.
	\item \textbf{Control of Support's Response Information:} Verify whether the Support uses any information from the Seeker's persona that has not been mentioned in the dialogue. To prevent the Support from making assumptions based on unmentioned information, we explicitly specify in the prompt,  ``\textit{Supporter should not mention information from `Seeker Persona Information' that seeker has not mentioned.}" This ensures coherence and role consistency throughout the dialogue.
	\item \textbf{Reasoning Validity of Support's Responses:} Examine the reasoning process of the Supporter, ensuring that the Seeker's situation, thoughts, and actions are accurately described, and that the strategies employed by the Supporter are reasonable.
	\item \textbf{Consistency between Support's Responses and Strategy:} Check whether the Supporter's responses align with the defined strategy and reflect the goal of the support strategy.
	\item \textbf{Overall Dialogue Structure Check:} Review the distribution of support strategies across the entire conversation to ensure it aligns with the three-phase framework of emotional support. This helps avoid an overreliance on a single strategy, making the Supporter appear more professional.
	\item \textbf{Dialogue Detail Review:} Conduct a thorough check of the dialogue in terms of the Seeker's level of information, the Supporter's understanding and helpfulness, the coherence of the conversation, and role consistency to ensure the fluency and consistency of the dialogue.
	\item \textbf{Format and Standardization Check:} Finally, ensure that the generated dialogue adheres to the proper format, especially verifying that the reasoning chain nodes are complete and that the overall content meets the specified format requirements.
\end{itemize}
We present a dialogue example in SSConv in Table \ref{tab:dialogue_example}. After dialogue generation, we conduct human quality evaluation following the guideline in Table~\ref{tab:guideline}.
\begin{table*}[h!]
\small
\centering
\resizebox{1.0\textwidth}{!}{
\begin{tabular}{|p{\textwidth}|}
\hline
This study aims to evaluate the quality of the conversation between a help-seeker (i.e., a person who is seeking support) and a help-supporter (i.e., a person who is providing support). Each task provides you with a single conversation. You need to evaluate the conversation from the following aspects.	\\ \\

%\\\textbf{$<$Supporter Evaluation$>$}\\
$<$\textbf{Informativeness}$>$ The help-seeker’s level of detail in describing his/her own emotional problems.\\
- \textbf{Score Options}: Detailed description (3), Moderate description (2), Minor description (1), No description (0)\\
- \textbf{Postive Example}: ``I feel so lonely after my dog passed away. He was my best friend'' provides a detailed description about both the seeker’s feelings and their situation. \\
- \textbf{Negative Example}:  ``I am sad'' does not tell us much about the seeker’s situation. For instance, the reason why they are sad is not mentioned. \\ \\

$<$\textbf{Understanding}$>$ The supporter’s grasp of the help-seeker’s experience and feelings. For example, how well the supporter understands and reflects on the help-seeker's emotions and situation.\ 

- \textbf{Score Options}: Deep understanding (3), Moderate understanding (2), Partial understanding (1), No understanding (0)\ \\
- \textbf{Positive Example}: Seeker: ``I’ve been feeling overwhelmed at work.'' Supporter: ``It sounds like work has been really stressful for you.'' This shows a clear understanding of the seeker’s emotions.\ \\
- \textbf{Negative Example}: Seeker: ``I’ve been feeling overwhelmed at work.'' Supporter: ``You should try to relax more.'' This response does not fully grasp the seeker’s feelings. \\ \\

$<$\textbf{Helpfulness}$>$ Whether the supporter is able to provide effective help and emotional support to the seeker (i.e., how much the supporter succeeds in making the seeker feel better and helping the seeker find a solution).\\
- \textbf{Score Options}: Completely helpful  (3), Mostly helpful  (2), A little helpful  (1), Not helpful (0)\\
- \textbf{Postive Example}: Seeker: ``I broke up, and I feel really sad." Supporter:``The pain of a breakup is really hard to bear, and I can feel your emotions right now. Can you tell me what was the most memorable part of this relationship for you? \\
- \textbf{Negative Example}:  Seeker: ``I broke up, and I feel really sad." Supporter:``Breakups are indeed painful, but you will definitely get through this tough time." \\ \\

$<$\textbf{Safety}$>$Whether the conversation contains unsafe contents.\\
- \textbf{Score Options}: Fully Safe (3), Mostly safe (2), Marginally safe (1), Unsafe (0)\\
- \textbf{Postive Example}: It does not involve toxic language, sensitive topics, immoral suggestions, etc. \\
- \textbf{Negative Example}: It involves toxic language, such as offensive content, hate speech, and biased opinions, or sensitive topics, such as medicine, violence, politics, gender, or race, or immoral suggestions (e.g., `cheating on your wife is a great idea').  \\ \\

$<$\textbf{Specificity}$>$ The level of detail in the help-seeker's emotional problem and the supporter's response. For example, a non-specific description/response refers to a general situation.\\
- \textbf{Score Options}: Highly specific  (3), Moderately specific  (2), Slightly specific (1), Not specific (0)\\
- \textbf{Postive Example}: ``I have been dating my boyfriend since high school, we've been together for ten years. Recently, …" This provides a detailed situation. \\
- \textbf{Negative Example}:  ``I broke up with my boyfriend" is a general description with no specific details.  \\ \\

$<$\textbf{Humanlikeness}$>$The degree to which the speaker sounds human.\\
- \textbf{Score Options}: Completely human-like. (3), Mostly human-like. (2), Somewhat human-like.  (1), Completely not human-like, obviously a machine. (0)\\
- \textbf{Postive Example}: ``You must be feeling really hurt, right? I can tell this has been an extremely heavy blow for you. Want to talk more about how you're feeling? I'm here to support you." \\
- \textbf{Negative Example}:  ``Your emotion has been recorded as negative. It is recommended to take step one: deep breathing. Step two:……" 
\\
\hline
\end{tabular}
}

\caption{Guideline of quality evaluation for emotional support conversations.}
\label{tab:guideline}
\end{table*}

% \begin{figure*}[h]
	%     \centering
	%     \includegraphics[width=0.60\linewidth]{figures/evalution_guidline.pdf}
	%     \caption{Guideline of human evaluation for dialogue quality.}
	%     \label{fig:dataset_evaluation_guidline}
	% \end{figure*}

%By following these stringent quality control processes, we significantly improved the overall quality of the LLM-generated dataset.

%\section{Detailed Data Statistics}
%\subsection{Topic and Subtopic}
%Our dataset encompasses 9 main topics, each containing multiple subtopics, resulting in over 100 subtopics in total. The distribution of topics is derived from the proportion of topics observed in real help-seeking scenarios. Since each dialogue may cover multiple subtopics, the proportions mentioned here refer to the percentage of subtopics relative to the total number of dialogues. Statistics of the Topic and Subtopics in SSconv are shown in Figure ~\ref{fig:subtopic}.
\

%\section{Guideline for Human Evaluation}

\section{Implementation Details}

\subsection{Data Preprocessing}

In the ESConv dataset, multiple consecutive utterances by the same speaker are recorded separately. To adhere to a strict turn-taking format, we preprocess the raw data by merging these consecutive utterances into single, cohesive statements, while preserving the strategies employed in each supporter utterance. For instance, a sequence of dialogues such as ``\textit{Seeker -u1, Seeker -u2, Supporter -(s3, u3), Supporter -(s4, u4)}'' is restructured into ``\textit{Seeker -(u1, u2), Supporter -(s3, s4, u3, u4)}''. Here, `\textit{s}' denotes strategy and `\textit{u}' indicates utterance. Additionally, we modify the dialogue to begin with the seeker and conclude with the supporter.

\subsection{Training Configurations}
We conduct experiments using one Tesla V100 GPU on LLaMA-Factory with the following training configurations:

\begin{itemize}
	\item Model Name: Llama-2-7b-hf
	\item Finetuning Type: LoRA
	\item LoRA Target: q\_proj, v\_proj
	\item Per-Device Train Batch Size: 2
	\item Gradient Accumulation Steps: 4
	\item Learning Rate Scheduler Type: Cosine
	\item Logging Steps: 10
	\item Learning Rate: 5e-5
	\item Number of Training Epochs: 5.0
	\item Plot Loss
	\item FP16: Enabled
	\item Cutoff Length: 10000
\end{itemize}

\end{document}